\documentclass[runningheads]{llncs}

\usepackage{eccv}

\usepackage{eccvabbrv}

\usepackage{graphicx}
\usepackage{booktabs}

\usepackage[accsupp]{axessibility}  

\usepackage{multirow}
\usepackage[table]{xcolor}
\usepackage{hyperref}

\usepackage{orcidlink}

\begin{document}

\title{SubSplat: High-Resolution Pixel-aligned 3DGS via Sub-pixel Gaussian Reparameterization} 

\titlerunning{SubSplat}

\author{Jiun Lee\inst{1}\orcidlink{0000-0002-7338-4565} \and
Jaekwang Kim\inst{2}\orcidlink{0000-0001-5174-0074} \and
Sangmin Lee\inst{3}\thanks{Corresponding author}\orcidlink{0000-0002-8606-2549}}

\authorrunning{Jiun Lee et al.}

\institute{AimFuture(c), \email{jiun.lee@aimfuture.ai}
  \and
  Sungkyunkwan University, \email{linux@skku.edu}
  \and
  Korea University, \email{sangmin-lee@korea.ac.kr}
}
\maketitle
\newcommand{\method}{SPGR}

\begin{abstract}
Pixel-aligned Gaussian splatting enables efficient and generalizable novel-view synthesis. However, high-resolution rendering faces a critical trade-off where increasing input resolution improves detail at the expense of quadratically rising network computational cost. Conversely, maintaining low-resolution inputs stabilizes this cost but results in insufficient Gaussian density and artifacts. To address this, we propose SubSplat, which introduces Sub-pixel Gaussian Reparameterizer(SPGR) to subdivide primary Gaussians into fine-grained primitives, restoring structural density directly from low-resolution features. We further enhance the reparameterization quality through feature aggregation, which effectively captures high-frequency details across multiple views. Experiments on RealEstate10K and ACID demonstrate that SubSplat achieves high-fidelity rendering with superior efficiency. Our results validate that the proposed framework successfully resolves the trade-off between reparameterization fidelity and network computational cost inherent in pixel-aligned Gaussian Splatting.
\end{abstract}

\section{Introduction}
\label{sec:intro}

Novel-View Synthesis (NVS) aims to generate photorealistic images of a scene from new camera viewpoints given a small set of calibrated input 
photos~\cite{mildenhall2021nerf,kerbl20233d,muller2022instant,chen2023mobilenerf,huang20242d}. NVS supports important applications in AR/VR telepresence, robotic perception, and content creation. For practical deployment, these applications require sharp, temporally stable views at interactive rates on high-resolution displays, including resource-constrained devices~\cite{kerbl20233d,muller2022instant,chen2023mobilenerf}. 
\begin{figure}[t!]
  \centering
  \includegraphics[width=0.98\linewidth]{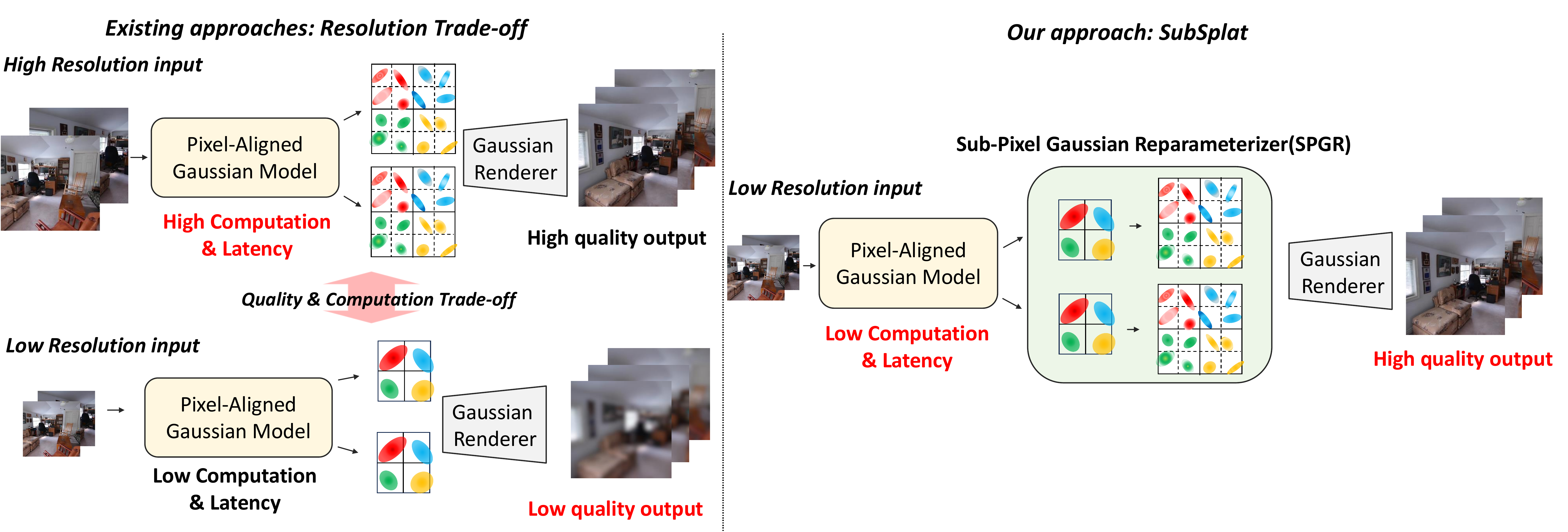} 
  
  \vspace{-5pt} 
  
  \caption{\textbf{Comparison of High-Resolution Rendering Strategies.} Unlike prior methods limited by fixed grids, SubSplat generates Sub-pixel Reparameterized Gaussians from low-resolution features to mitigate the quadratic cost of high-resolution inputs. This enables high-fidelity reconstruction at extended scales while maintaining stable network latency.}
  
  \label{fig:overview}
  \vspace{-10pt} 
\end{figure}

Recent neural rendering~\cite{chen2023mobilenerf,muller2022instant,kerbl20233d,barron2021mipnerf} advances have significantly improved NVS quality and practicality. These advances motivate designs that preserve image quality while keeping computation and memory costs predictable across devices and display scales. Among these advances, 3D Gaussian Splatting(3DGS)
~\cite{kerbl20233d} achieves high-quality rendering with learnable 3D anisotropic Gaussians and a differentiable rasterizer. Prior works improve rendering quality through anti-aliasing~\cite{yu2024mip,yan2024multi} and adaptive 
densification~\cite{kerbl20233d,rota2024revising,zhang2024pixel}. However, these methods require per-scene optimization, limiting generalization and preventing efficient deployment across diverse scenes. To tackle this problem, pixel-aligned Gaussian methods~\cite{charatan2024pixelsplat,chen2024mvsplat,xu2025depthsplat,zhang2025transplat,tang2024hisplat,zhang2024gs} train feed-forward networks across diverse scenes, enabling generalized Gaussian generation without per-scene optimization. These methods predict Gaussian primitives on a fixed image grid from sparse input views and render them with a differentiable splatting backend, achieving faster inference on novel scenes.

Despite recent progress, pixel-aligned methods face significant challenges in scaling to high-resolution rendering. These methods predict Gaussian attributes based on a fixed image grid, where the total number of primitives is intrinsically tied to the input resolution. This grid-based design creates a fundamental quality-efficiency dilemma, as the total system latency is primarily dominated by the backbone network's forward pass.

The straightforward strategy to achieve high-resolution rendering is to increase the input resolution accordingly. However, this causes quadratic computational growth in the backbone network, as the network must process a much denser grid to predict the additional primitives. Since the total latency is primarily dominated by this backbone forward pass, doubling the resolution requires $4\times$ more computation, making it impractical for interactive use.
To avoid this burden, one might use low-resolution inputs and render the resulting sparse Gaussians on a high-resolution screen. Yet, this leads to insufficient Gaussian density relative to the target pixels. This under-parameterization inevitably results in noticeable blur and halo artifacts, as the limited number of grid-anchored primitives cannot adequately represent high-frequency details or sharp structural edges. Consequently, existing pixel-aligned methods are trapped in a trade-off: they must choose between (i) quadratic increase in backbone computational cost when the input resolution is scaled up to match the target output resolution or (ii) quality degradation when forcing high-resolution outputs from coarse input grids.

To address the computational inefficiency and quality degradation of existing pixel-aligned models, we propose SubSplat, a novel framework featuring Sub-pixel Gaussian Reparameterizer(SPGR). As shown in Fig.~\ref{fig:overview}, our approach moves beyond the constraints of fixed grid-sampling by introducing SPGR that redistributes Gaussian geometry and appearance for the target resolution. Instead of increasing the backbone’s input resolution, which leads to a quadratic increase in computational cost, our method maintains a stable backbone latency by utilizing low-resolution inputs. The reparameterizer transforms each grid-anchored primary Gaussian into a set of fine-grained sub-pixel primitives through an efficient module. This enables content-aware density scaling to higher target resolutions (e.g. $\times 2$, $\times4$ the input) with negligible computational overhead, effectively restoring structural details while preserving inference efficiency.

To further enhance these sub-pixel primitives, we incorporate deformable attention to align  Gaussian primitives with multi-view features. This mechanism aggregates contextual cues across views, ensuring that the increased Gaussian density translates into sharp details and geometric consistency. Together, these designs allow our method to produce high-fidelity, high-resolution outputs while maintaining a highly efficient inference pipeline. By separating the output primitive density from the backbone’s computational resolution, we achieve high-resolution rendering without a quadratic increase in network costs. This approach effectively solves the structural bottleneck where high-quality results previously depended on using heavy and expensive backbone grids.
\\\\
The contributions of this work are as follows:
\begin{itemize}
\item We propose SubSplat, a novel framework that enables high-resolution rendering while keeping the input resolution low, preventing the quadratic scaling of network computation costs which exist in the previous pixel-aligned approaches.
\item We design a Sub-pixel Gaussian Reparameterizer(SPGR) that subdivides each grid-anchored Gaussian into fine-grained primitives through dedicated geometry and appearance features. Additionally, a footprint-aware opacity redistribution mechanism ensures that total opacity is conserved across subdivisions.
\item SubSplat achieves state-of-the-art performance across multiple target scales, delivering over 3$\times$ latency reduction compared to full-resolution baselines while providing superior reconstruction quality that surpasses both pixel-aligned methods and even external upsampling approaches.
\end{itemize}

\section{Related Work}
\label{sec:related}

\noindent\textbf{3D Gaussian splatting.}
3DGS~\cite{kerbl20233d} represents scenes with anisotropic Gaussians and a differentiable rasterizer, enabling real-time novel-view synthesis and fast optimization. 
Early point-based renderers~\cite{rusinkiewicz2000qsplat,botsch2003high} pioneered point-based representations with multi-resolution hierarchies.
EWA splatting for points and surfaces~\cite{zwicker2001surface,zwicker2002ewa} established footprint-based anti-aliasing.
Building on this foundation, Mip-Splatting~\cite{yu2024mip}, Multi-Scale 3DGS~\cite{yan2024multi}, Analytic-Splatting~\cite{liang2024analytic}, and Anti-Aliased 2DGS~\cite{younes2026anti} based on ~\cite{huang20242d} further mitigate aliasing across scales. Adaptive densification adds Gaussians in regions with large gradients~\cite{kerbl20233d}, with recent work refining schedules and density control~\cite{rota2024revising,zhang2024pixel,scaffoldgs,fang2024mini}. 
However, these approaches still rely on per-scene optimization and fixed resolutions, limiting their generalization and scalability. To address this, SubSplat employs a feed-forward architecture with a subdivision mechanism to mitigate the resolution constraints of the feature grid. This allows for fine-grained reconstruction while reducing the dependency on the initial feature scale.

\noindent\textbf{High-resolution novel-view synthesis(NVS).}
High-resolution NVS typically relies on per-scene optimization with various supervision strategies. NeRF ~\cite{mildenhall2021nerf}-based methods~\cite{wang2022nerf,huang2023refsr,lin2024fastsr} enhance sub-pixel details but require high-resolution ground truth or suffer from slow rendering. Recent approaches leverage external 2D priors: single-image SR 
models~\cite{yoon2023cross} or diffusion models with 3D 
synchronization~\cite{lee2024disr}. In the 3DGS line, 
GaussianSR~\cite{yu2024gaussiansr} applies diffusion-based Score Distillation Sampling, SuperGS~\cite{xie2024supergs} introduces a coarse-to-fine framework with latent feature fields, and S2Gaussian~\cite{wan2025s2gaussian} densifies low-resolution Gaussians through Shuffle Split and refines with super-resolved multi-view supervision. SRGS~\cite{feng2024srgs} uses pre-trained SR 
backbones~\cite{liang2021swinir} to provide pseudo-HR supervision, though multi-view inconsistencies can introduce ambiguities. While effective at enhancing detail, these methods require per-scene optimization, limiting cross-scene generalization. Pixel-aligned feed-forward pipelines address this by enabling immediate inference without scene-specific tuning.

\noindent\textbf{Pixel-aligned Gaussian splatting.}
Pixel-aligned methods~\cite{charatan2024pixelsplat,chen2024mvsplat,tang2024hisplat,xu2025depthsplat,zhang2025transplat}
train feed-forward networks to predict Gaussian parameters on a fixed image grid, enabling generalized novel-view synthesis without per-scene optimization.
Multi-view variants~\cite{chen2024mvsplat,tang2024hisplat} aggregate information across views via cross-attention and construct plane-sweep cost volumes to guide correspondence and geometry hypotheses~\cite{collins1996space,yao2018mvsnet,xu2023unifying}. By relying on a fixed number of Gaussians tied to the input grid, these designs face an inherent quality-cost constraint. Improving reconstruction detail necessitates higher input resolutions, which triggers a quadratic surge in computational overhead.
Depth-guided variants~\cite{xu2025depthsplat,zhang2025transplat} leverage pre-trained depth priors to improve reconstruction quality, yet remain grid-anchored at a fixed feature scale.
Recent refinement approaches~\cite{fei2024pixelgaussian,nam2025generative} add detail by modulating per-pixel attributes or synthesizing additional Gaussians.
However, these methods remain constrained by the feature grid resolution, making output fidelity dependent on the backbone's computational scale. To address this, SubSplat introduces a subdivision mechanism that generates sub-pixel primitives from coarse Gaussians. This alleviates the resolution dependency, enabling high-fidelity rendering with improved computational efficiency.

\begin{figure*}[t]
  \centering
  \includegraphics[width=\textwidth]{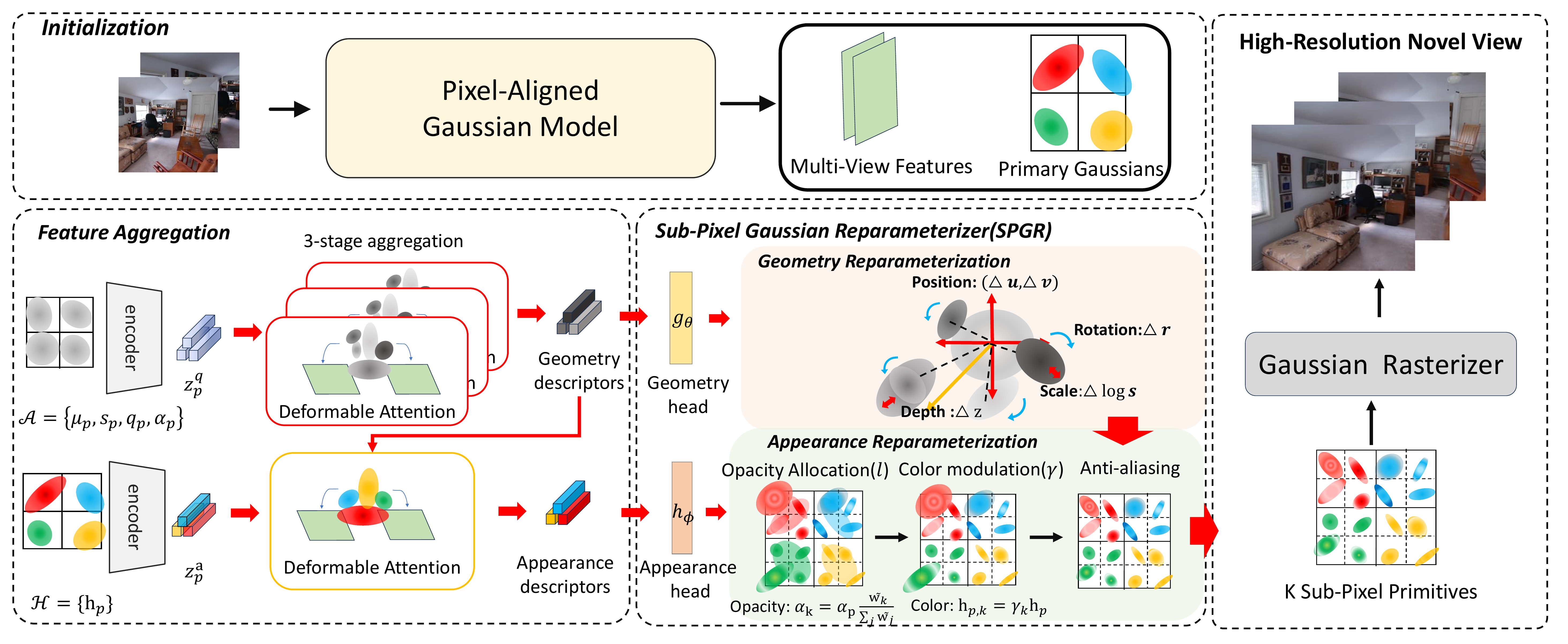}
\caption{\textbf{Overview of the SubSplat Architecture.} SubSplat aggregates multi-view features via Deformable Attention to extract rich geometry and appearance descriptors. These descriptors guide the Sub-pixel Gaussian Reparameterizer(SPGR) to subdivide Gaussians into fine-grained primitives, restoring high-resolution details while incurring only a marginal increase in total inference time.}
  \label{fig:subpixel}
\end{figure*}

\section{Proposed Method} 
\label{sec:Proposed}

\subsection{Overview}
SubSplat is an end-to-end pipeline that synthesizes high-resolution scenes from coarse feature representations. The process begins with a pixel-aligned backbone~\cite{chen2024mvsplat} initializing primary Gaussians. To enrich these primitives, a deformable-attention module~\cite{huang2024gaussianformer,fei2024pixelgaussian} performs feature aggregation to integrate auxiliary geometry and appearance information, providing the necessary contextual cues for the reparameterization stage. At the core of our method, a sub-pixel Gaussian reparameterizer(SPGR) subdivides these primary Gaussians into multiple sub-pixel primitives. This subdivision alleviates the resolution dependency on the backbone's feature scale, while an attribute redistribution strategy modulates color and preserves integrated opacity for high-fidelity reconstruction. By shifting detail enhancement to this subdivision stage, SubSplat recovers sharp edges and fine textures with only a marginal increase in network computational overhead.

\subsection{Feature Aggregation}

\subsubsection{Initialization}
We use the same input structure as MVSplat~\cite{chen2024mvsplat}, keeping camera metadata separate from learnable states.

\noindent\textbf{Anchors and color.} 
From the backbone output, let $N$ primary Gaussians be
\begin{equation}
\mathcal{A} = \{\mathbf{a}_p\}_{p=1}^{N}, \qquad
\mathbf{a}_p = [\,\boldsymbol{\mu}_p, \mathbf{s}_p, q_p, \alpha_p\,] \in \mathbb{R}^{11},
\end{equation}
together with their corresponding spherical‑harmonic(SH) color
\begin{equation}
\mathcal{H} = \{\mathbf{h}_p \in \mathbb{R}^{C \times D_{\mathrm{sh}}}\}_{p=1}^{N}.
\end{equation}
Here $\mathbf \mu_p$ denotes the world-space position, and $\mathbf s_p$ are the per-axis scales that define the anisotropic footprint of the Gaussian in its local frame. The unit quaternion $\mathbf q_p$ represents its orientation. We use degree‑4 SH by default ($C{=}3$, $D_{\mathrm{sh}}{=}25$). These sets provide the anchor states and colors that downstream modules will refine.\\
\noindent\textbf{Backbone features.}
We also take multi‑view feature maps
\begin{equation}
\mathcal{F} = \{\Phi_{v,\ell}\}_{v=1..V,\ \ell=1..L}, \qquad
\Phi_{v,\ell} \in \mathbb{R}^{H_\ell \times W_\ell \times D_\ell},
\end{equation}
which will be sampled by deformable attention in the feature aggregation stage.

\noindent\textbf{Camera metadata.}
Camera intrinsics and extrinsics are fixed and provide the projection information 
used in both the deformable-attention feature aggregation and the SPGR.

\subsubsection{Geometry feature aggregation}
Given primary Gaussian states $\mathcal A$ and multi-view features $\mathcal F$, we form a per-anchor geometry query as a sum of attribute-wise embeddings followed by a projection using an MLP:
\begin{equation}
\mathbf{z}^{g}_p = 
W_g\!\big(
\phi_x(\boldsymbol{\mu}_p) +
\phi_s(\mathbf{s}_p) +
\phi_q(q_p) +
\phi_\alpha(\alpha_p)
\big).
\end{equation}
Each Gaussian then aggregates multi-view evidence with a deformable-attention operator 
over three aggregation stages:
\begin{equation}
\begin{aligned}
\mathbf{f}^{g,0}_p &= \mathbf{0},\\
\mathbf{f}^{g,t+1}_p &= \mathrm{DA}\!\big(\mathcal{F};\, \mathbf{z}^{g}_p,
\mathbf{f}^{g,t}_p, \mathcal{P}(\boldsymbol{\mu}_p), \mathbf{a}_p\big),
\quad t = 0, 1, 2,\\
\mathbf{f}^{g}_p &= \mathbf{f}^{g,3}_p.
\end{aligned}
\end{equation}
Here, $\phi_x$, $\phi_s$, $\phi_q$, and $\phi_\alpha$ are small per-attribute MLPs, 
and $W_g$ is a linear projection. 
We align the geometry embedding dimension with the multi-view feature space 
and compute view-dependent projection matrices from the camera metadata. 
Each primary Gaussian is projected onto the reference views through these matrices, 
and the resulting 2D projections are used by the deformable attention module~\cite{huang2024gaussianformer}  
to aggregate view-specific geometric context.
Inspired by Deformable DETR’s deformable attention~\cite{zhu2020deformable}, we adopt a multi-stage design and stack three deformable-attention layers to progressively aggregate cross-view features.
This multi-stage design enhances geometric consistency by integrating local-to-global 
context across multiple feature scales.
 This process yields the geometry descriptor \(\mathbf f^{g}_p\), a compact, view-consistent representation of cross-view geometry for subsequent SPGR.

\subsubsection{Appearance feature aggregation}
We map the primary SH coefficients to an appearance query:
\begin{equation}
\mathbf{z}^{a}_p = \phi_c\!\big(\mathrm{vec}(\mathbf{h}_p)\big),
\end{equation}
where \(\phi_c\) is an MLP that produces an embedding aligned with the 
multi-view feature dimension, and \(\mathrm{vec}(\mathbf h_p) \in \mathbb{R}^{C \cdot D_{\mathrm{sh}}}\) 
flattens the spherical-harmonic coefficients. We then aggregate appearance 
features via deformable attention, conditioned on the geometry descriptor 
\begin{equation}
\mathbf{f}^{a}_p = 
\mathrm{DA}\!\big(
\mathcal{F};\, \mathbf{z}^{a}_p,
\mathbf{f}^{g}_p,
\mathcal{P}(\boldsymbol{\mu}_p),
\mathbf{a}_p
\big).
\end{equation}
The geometry descriptor \(\mathbf f^{g}_p\) serves as the conditioning input, 
guiding the aggregation based on the established geometric context. By reusing 
the same projection context \(\mathcal P(\mathbf \mu_p)\) and primary anchor 
\(\mathbf a_p\) as the geometry path, this design allows geometry to determine 
\textit{where to look} and appearance to determine \textit{what to amplify}, 
maintaining structural integrity and improving color consistency across multiple views.

\subsection{Sub-Pixel Gaussian Reparameterizer}
Each primary Gaussian $p$ is associated with geometry and appearance descriptors $\mathbf{f}^{g}_p, \mathbf{f}^{a}_p \in \mathbb{R}^{E}$, where $E$ denotes the shared embedding dimension of the feature space used by the prediction heads. For each primary Gaussian $p$, the Sub-pixel Gaussian Reparameterizer(SPGR) divides it into $K$ sub-pixel primitives at the target resolution. Here, $K$ is defined as the squared ratio between the target and input resolutions (e.g., $K=4$ for a $2\times$ resolution scale and $K=16$ for a $4\times$ resolution scale). Using the aggregated descriptors $\mathbf{f}^{g}_p$ and $\mathbf{f}^{a}_p$, two dedicated prediction heads generate primitive-level refinements. Specifically, a geometry head $g_\theta$ predicts spatial properties including position, depth, scale, and rotation, while an appearance head $h_\phi$ predicts color and opacity adjustments. These primitive-level attributes are composed with the primary Gaussian's properties to construct the final primitives. Each process is described in detail below.

\subsubsection{Geometry Reparameterization}

The geometry head $g_\theta: \mathbb{R}^E \to \mathbb{R}^{K \times 9}$ predicts spatial properties for each sub-pixel primitive:
\begin{equation}
[\Delta u_k, \Delta v_k, \Delta z_k, 
\Delta\!\log\mathbf{s}_k, \Delta\mathbf{r}_k]_{k=1}^{K} 
= g_\theta(\mathbf{f}^{g}_p).
\end{equation}
Here, $(\Delta u_k, \Delta v_k) \in \mathbb{R}^2$ are the sub-pixel offsets within the primary Gaussian's footprint, $\Delta z_k \in \mathbb{R}$ is the depth residual, $\Delta\log\mathbf{s}_k \in \mathbb{R}^3$ denotes the log-scale residuals for anisotropic scaling, and $\Delta \mathbf{r}_k \in \mathbb{R}^3$ represents the axis-angle rotation.

\noindent\textbf{Position and depth.}
We select, for each anchor, as the primary view the camera in which the anchor is visible and has the smallest positive depth~\cite{yao2018mvsnet,chen2024lara}. 
SPGR then predicts sub-pixel Gaussian position and depth in that view. Specifically, $K$ sub-Gaussians are anchored to a uniform grid centered on the primary Gaussian.
Let $z_{\mathrm{sel}}$ denote this depth. Each sub-pixel offset $(\Delta u_k,\Delta v_k)$ is back-projected through the intrinsics and adjusted along the viewing ray~\cite{hartley2003multiple,collins1996space}:
\begin{equation}
\mathbf{\mu}_{p,k}
= \mathbf{\mu}_{p}
+ R_{c2w}
\begin{bmatrix}
z_{\mathrm{sel}}\Delta u_k / f_x\\
z_{\mathrm{sel}}\Delta v_k / f_y\\
0
\end{bmatrix}
+ \Delta \hat{z}_k\, R_{c2w}\!\begin{bmatrix}0\\0\\1\end{bmatrix},
\end{equation}
where $(f_x,f_y)$ are the focal lengths and $R_{c2w}$ is the camera-to-world rotation.
To keep the reparameterization stable across resolutions and viewpoints, the depth residual is bounded via a $\tanh$ mapping,
\begin{equation}
\Delta \hat{z}_k = \tanh(\Delta z_k),
\end{equation}
which constrains updates along the ray within a small, scene-independent range. Please refer to Supplementary~\hyperref[sec:supp_A3]{A.3} for more details.

\noindent\textbf{Anisotropic scale and rotation.}
Each sub-pixel Gaussian inherits its primary Gaussian’s orientation and scale, refined locally 
in the original coordinate system~\cite{kerbl20233d}. 
The updated scale is computed as
\begin{equation}
\mathbf{s}_{p,k} = \mathbf{s}_p \odot \exp(\Delta\!\log\mathbf{s}_k).
\end{equation}
where $\odot$ denotes element-wise multiplication, allowing each sub-pixel Gaussian to stretch 
or compress independently along the three axes while remaining numerically stable.

For rotation, the residual $\Delta \mathbf r_k \in \mathbb{R}^3$ is an axis–angle vector 
(in radians) that represents a small local rotation~\cite{sola2017quaternion} with respect to the sub-pixel Gaussian's orientation. 
This residual is converted into a quaternion, composed with the sub-pixel Gaussian's quaternion, 
and then normalized to form the final orientation:
\begin{equation}
q_{p,k} = \mathrm{norm}\!\big(q_p \otimes q(\Delta \mathbf{r}_k)\big).
\end{equation}
Following 3D Gaussian Splatting~\cite{kerbl20233d}, 
the rotation matrix $R(q_{p,k})$ and scale matrix 
$\mathrm{diag}(\mathbf s_{p,k}^2)$ are combined to form the covariance:
\begin{equation}
\Sigma_{p,k} = R(q_{p,k})\,\mathrm{diag}(\mathbf{s}_{p,k}^2)\,R(q_{p,k})^\top.
\end{equation}

\subsubsection{Appearance Reparameterization}
The appearance head $h_\phi: \mathbb{R}^E \to \mathbb{R}^{K \times 2}$ 
predicts two key components for each sub-pixel primitive:
\begin{equation}
[\ell_k, \gamma_k]_{k=1}^{K} = h_\phi(\mathbf{f}^{a}_p).
\end{equation}
The first term $\ell_k$ is a distribution logit that partitions the primary Gaussian's opacity among its primitives via softmax normalization, ensuring the integrated opacity is preserved. 
The second term $\gamma_k$ is a bounded scalar for color modulation, which refines the primary Gaussian's color to capture sub-pixel appearance details. Together, they achieve high-fidelity reconstruction by shifting detail enhancement to the subdivision stage.

\noindent\textbf{Opacity allocation.}
To maintain integrated opacity, we distribute the primary Gaussian's opacity $\alpha_p$ among its assigned sub-pixel primitives. This process is guided by a footprint-aware formulation~\cite{liang2024analytic,younes2026anti}, concentrating opacity on smaller primitives to sharpen structural boundaries. By giving higher weights to primitives with smaller screen-space footprints, the model effectively sharpens structural edges. A final renormalization ensures that the total opacity of the primitives remains equal to that of the primary source, satisfying $\sum_k \alpha_k = \alpha_p$.
Let $J_p \in \mathbb{R}^{2\times 3}$ be the projection Jacobian at the primary position. The projected area of each primitive is estimated as:
\begin{equation}
A_k \approx \sqrt{\det\bigl(J_p,\Sigma_{k},J_p^\top\bigr)}.
\end{equation}
We derive the inverse-area weights by combining the distribution logits $\ell_k$ with the estimated footprints:\begin{equation}w_k = \mathrm{softmax}(\ell_k), \qquad \tilde{w}_k = \frac{w_k}{A_k+\varepsilon}.\end{equation}The final opacity for each primitive is then assigned as:\begin{equation}\alpha_k = \alpha_p \cdot \frac{\tilde{w}_k}{\sum_j \tilde{w}_j}.\end{equation}This mechanism enforces distributional integrity from the primary Gaussian while stabilizing high-resolution synthesis by prioritizing density on intricate scene details.

\noindent\textbf{Color modulation.}
The color of each sub-pixel Gaussian is scaled by a bounded gain factor $\gamma_k$:
\begin{equation}
\mathbf{h}_{p,k} = \gamma_k \, \mathbf{h}_p.
\end{equation}
Here, $\mathbf{h}_p \in \mathbb{R}^{C \times D_{\mathrm{sh}}}$ denotes the primary Gaussian's spherical harmonic coefficients. This operation modulates color intensity within a stable range while preserving the primary color direction, consistent with multiplicative adjustments used for color calibration in Gaussian-based rendering~\cite{rematas2022urban,kulhanek2024wildgaussians}. By scaling the existing coefficients, the model introduces fine-grained appearance variations without deviating from the color context inherited from the primary Gaussian.

\noindent\textbf{Anti-aliasing.} From an anti-aliasing perspective, $A_k$ represents the projected screen-space footprint of each primitive. To stabilize sub-pixel Gaussian rendering, we enforce a minimum area $A_{\min}$; when $A_k < A_{\min}$, the in-plane scales are minimally adjusted to ensure consistent pixel coverage. Consistent with scale-aware formulations for 3DGS~\cite{liang2024analytic,younes2026anti}, this clamping strategy preserves the effective footprint, mitigating flickering and aliasing artifacts during high-resolution synthesis. By maintaining a lower bound on the primitive size, the model ensures that the reparameterization stage remains robust even when generating extremely fine-grained structures.

\subsection{Training Objective}
Following MVSplat~\cite{chen2024mvsplat}, 
our model predicts sub-pixel 3D Gaussian parameters and renders novel views from them. 
The network is trained with ground-truth RGB images using a weighted combination of $\ell_2$ and LPIPS~\cite{zhang2018unreasonable}$^{\!}$ losses:
\begin{equation}
\mathcal{L} = \mathcal{L}_{\ell_2} + \lambda\,\mathcal{L}_{\mathrm{LPIPS}},
\end{equation}
where $\lambda$ is the weighting hyperparameter that balances the two losses. We set $\lambda$ to 0.05 following the common protocol \cite{chen2024mvsplat,charatan2024pixelsplat,zhang2025transplat}.

\section{Experiments}
\label{sec:exp}
To evaluate SubSplat, we conduct comparative experiments focusing on the trade-off between rendering resolution and computational cost. First, we reconstruct scenes from $256\times 256$ inputs and render them at $512\times 512$ ($\times 2$) and $1024\times 1024$ ($\times 4$) resolutions to observe structural details. We compare our model against baselines where the input and output resolutions are coupled at $512\times512$. This analysis validates that SubSplat achieves comparable fidelity while significantly reducing network overhead, specifically in inference latency and peak memory. We also contrast our 3D primitive-level refinement with image-space upsamplers, such as bilinear interpolation and HiT-SR~\cite{zhang2024hit}, to assess the impact of densifying geometry in 3D space. Finally, we perform an ablation study to verify the effectiveness of each component in our model for scene reconstruction.

\begin{table*}[t!]
\captionsetup{position=top}
\caption{Quantitative evaluation of resolution scalability on RealEstate10K~\cite{zhou2018stereo} and ACID~\cite{liu2021infinite}. All methods reconstruct scenes from a fixed $256 \times 256$ input resolution, and performance is evaluated at $\times 2$ ($512 \times 512$) and $\times 4$ ($1024 \times 1024$) output scales to assess high-fidelity reconstruction capabilities. Metrics: PSNR$\uparrow$ / SSIM$\uparrow$ / LPIPS$\downarrow$. Best and second-best results are bold and \underline{underlined}, respectively.}
\label{tab:re10k-acid-main}
\centering
\setlength{\tabcolsep}{3pt}
\renewcommand{\arraystretch}{1.15}
\resizebox{\textwidth}{!}{
\begin{tabular}{l ccc ccc ccc ccc}
\toprule
\multirow{2}{*}{Method} &
\multicolumn{6}{c}{\textbf{RealEstate10K}~\cite{zhou2018stereo}} &
\multicolumn{6}{c}{\textbf{ACID}~\cite{liu2021infinite}} \\
\cmidrule(lr){2-7}\cmidrule(lr){8-13}
& \multicolumn{3}{c}{Output($\times 2$) $512 \times 512$} & \multicolumn{3}{c}{Output($\times 4$) $1024\times 1024$}
& \multicolumn{3}{c}{Output($\times 2$) $512 \times 512$} & \multicolumn{3}{c}{Output($\times 4$) $1024 \times 1024$} \\
\cmidrule(lr){2-4}\cmidrule(lr){5-7}\cmidrule(lr){8-10}\cmidrule(lr){11-13}
& PSNR$\uparrow$ & SSIM$\uparrow$ & LPIPS$\downarrow$
& PSNR$\uparrow$ & SSIM$\uparrow$ & LPIPS$\downarrow$
& PSNR$\uparrow$ & SSIM$\uparrow$ & LPIPS$\downarrow$
& PSNR$\uparrow$ & SSIM$\uparrow$ & LPIPS$\downarrow$ \\
\midrule
PixelSplat~\cite{charatan2024pixelsplat} & \underline{23.44} & 0.789 & 0.227 & 21.44 & 0.767 & 0.323 & \underline{24.41} & 0.709 & 0.292 & \underline{22.95} & 0.661 & 0.409 \\
MVSplat~\cite{chen2024mvsplat}           & 19.46 & 0.751 & 0.269 & 16.98 & 0.671 & 0.424 & 18.38 & 0.645 & 0.349 & 16.57 & 0.616 & 0.489 \\
TranSplat~\cite{zhang2025transplat}      & 19.61 & 0.756 & 0.265 & 17.16 & 0.679 & 0.424 & 18.64 & 0.657 & 0.333 & 16.93 & 0.642 & 0.499 \\
DepthSplat~\cite{xu2025depthsplat}       & 19.54 & 0.720 & 0.309 & 16.29 & 0.587 & 0.461 & 17.19 & 0.535 & 0.469 & 16.32 & 0.630 & 0.476 \\
HiSplat~\cite{tang2024hisplat} & 23.26 & \underline{0.803} & \underline{0.207} & \underline{22.05} & \underline{0.779} & \underline{0.309} & 23.59 & \underline{0.724} & \underline{0.273} & 22.03 & \underline{0.674} & \underline{0.388} \\
\rowcolor{gray!15}
\textbf{Ours (SubSplat)}                 & \textbf{25.52} & \textbf{0.850} & \textbf{0.167} & \textbf{22.65} & \textbf{0.781} & \textbf{0.268} & \textbf{26.03} & \textbf{0.775} & \textbf{0.216} & \textbf{23.35} & \textbf{0.674} & \textbf{0.330} \\
\bottomrule
\end{tabular}}
\end{table*}

\subsection{Experimental Setup}

\noindent\textbf{Datasets.} All data follows the PixelSplat mining procedure~\cite{charatan2024pixelsplat} to ensure consistency across evaluations. Following the MVSplat setup, SubSplat and all baselines evaluate sparse-view performance using exactly two input views with frame gaps of 45--192. We train all models on the 720p RealEstate10K~\cite{zhou2018stereo} training split, which consists of 64,002 scenes. During inference, the 720p Real-Estate10K ~\cite{zhou2018stereo} test set (7,064 scenes) is used to evaluate $\times 2$ resolution scalability, backbone efficiency, and comparisons with image-space upsamplers, as well as for the ablation study. We utilize the 1080p RealEstate10K split for $\times 4$ inference evaluations and ablation studies. Due to the scarcity of high-resolution training data(1080p), we instead train a base model on 720p sequences using $128 \times 128$ inputs to target $512 \times 512$ outputs. This configuration is then directly applied to $1024 \times 1024$ inference from $256 \times 256$ inputs, validating SubSplat’s resolution scalability across 3,703 high-resolution scenes.
To evaluate cross-dataset generalization, we test the models on the 720p ACID~\cite{liu2021infinite} dataset (2,026 scenes) for $\times 2$ scene reconstruction. Furthermore, we assess $\times 4$ generalization using the 1080p ACID~\cite{liu2021infinite} dataset, which contains 1,622 test scenes.

\noindent\textbf{Baselines.} We compare SubSplat with pixel-aligned models, PixelSplat~\cite{charatan2024pixelsplat}, MVSplat~\cite{chen2024mvsplat} and HiSplat~\cite{tang2024hisplat}, and depth-guided variants, TranSplat~\cite{zhang2025transplat} and DepthSplat~\cite{xu2025depthsplat}. To evaluate the benefits of 3D-level densification, we also compare against image-space upsamplers, including bilinear interpolation and Hit-SR~\cite{zhang2024hit}. These upsamplers are applied to MVSplat~\cite{chen2024mvsplat} by reconstructing scenes from $256 \times 256$ inputs and rendering at $512 \times 512$ resolution.

\noindent\textbf{Implementation Details.} SubSplat is trained for 300K iterations on a single NVIDIA A100 (80 GB) GPU with a batch size of 12. We use the Adam optimizer with a learning rate of $2 \times 10^{-4}$ and a OneCycleLR scheduler featuring a 2K-step warm-up. Our primary model is trained to reconstruct scenes from $256 \times 256$ inputs and render them at $512 \times 512$ resolution. To analyze resolution scalability for extensive output resolutions, we train a model on a $128 \times 128$ to $512 \times 512$ task ($\times 4$ scaling). This trained model is then directly applied to evaluate $1024 \times 1024$ inference from $256 \times 256$ inputs, demonstrating its ability to maintain high-fidelity performance across larger absolute scales while preserving the same $\times 4$ scaling ratio. Efficiency is evaluated by comparing our $256 \times 256$ input model against baselines using $512 \times 512$ inputs. All efficiency metrics are measured on the same NVIDIA A100 GPU to ensure a fair comparison. Regarding the inference setup, baselines are not evaluated under the $1024 \times 1024$ input setting because their memory-intensive architectures result in Out-of-Memory (OOM) issues. All image-space upsamplers, including HiT-SR~\cite{zhang2024hit}, are applied to MVSplat~\cite{chen2024mvsplat} outputs using the $256 \times 256$ to $512 \times 512$ configuration.
All variants in the ablation study, including the baseline, were trained using the exact same objective with low-resolution inputs and high-resolution target.

\begin{figure*}[t!]
\centering
\includegraphics[width=\textwidth]{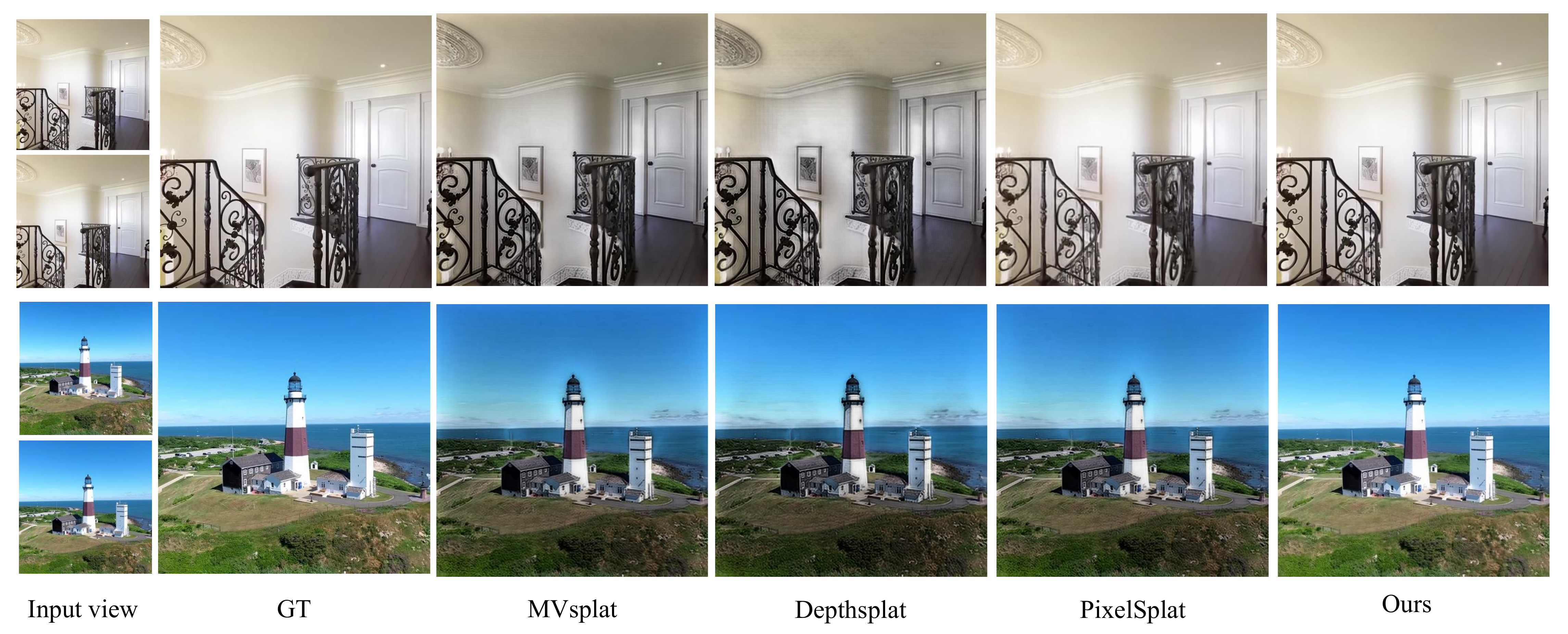}
\vspace{-20pt}
\caption{
\textbf{Qualitative results at $512 \times 512$ screen space.} 
While pixel-aligned baselines suffer from geometric instability and blurring when rendering beyond their $256 \times 256$ input grid, SubSplat(ours) recovers sharp, structurally consistent details on RealEstate10K~\cite{zhou2018stereo} and ACID~\cite{liu2021infinite} through sub-pixel Gaussian reparameterization.
}
\label{fig:qual_256_512}
\vspace{-10pt}
\end{figure*}
\begin{figure*}[t]
\centering
\includegraphics[width=\textwidth]{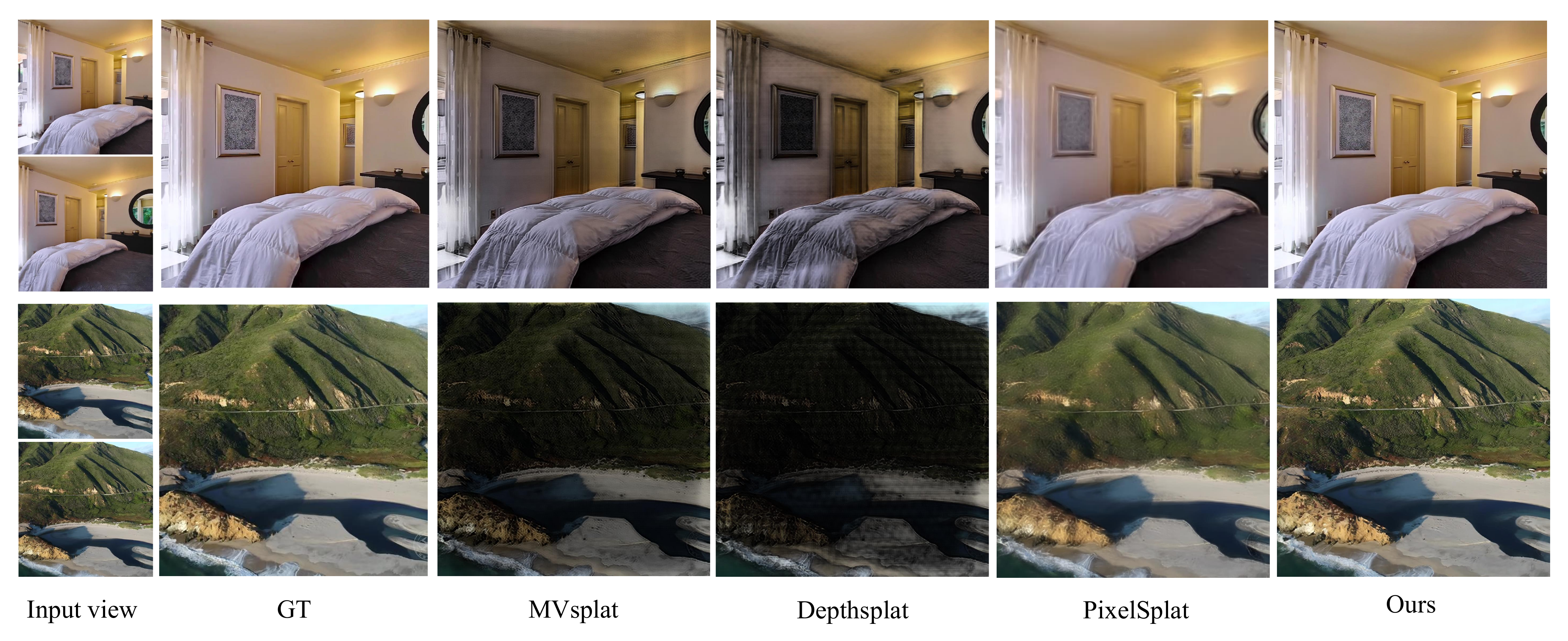}
\vspace{-20pt}
\caption{
\textbf{Qualitative results at $1024 \times 1024$ screen space.} 
Even with a $256 \times 256$ input, SubSplat(ours) maintains structural consistency and sharp details on RealEstate10K~\cite{zhou2018stereo} and ACID~\cite{liu2021infinite}. In contrast, pixel-aligned baselines exhibit blurring or severe halos when rendering at $4\times$ resolution.
}
\vspace{-10pt}

\label{fig:qual_1024}

\end{figure*}

\subsection{Results and Analysis}

\noindent\textbf{Resolution Scalability.} 
Figures~\ref{fig:qual_256_512} and~\ref{fig:qual_1024} demonstrate that SubSplat recovers sharper geometry and more stable textures than existing models across various output scales. While MVSplat~\cite{chen2024mvsplat} and DepthSplat~\cite{xu2025depthsplat} suffer from edge flickering and aliasing at $512 \times 512$ and $1024 \times 1024$, SubSplat maintains structural clarity and fine details through its sub-pixel Gaussian reparameterization. Interestingly, PixelSplat~\cite{charatan2024pixelsplat} shows relatively less fidelity loss than other baselines by assigning a fixed number of three Gaussians per pixel, providing better coverage for high-resolution grids. Nevertheless, SubSplat consistently outperforms PixelSplat~\cite{charatan2024pixelsplat} by exhibiting superior robustness to increasing output scales, achieved by directly optimizing high-density Gaussian primitives for multi-scale rendering. Quantitative results in Table~\ref{tab:re10k-acid-main} further validate this robustness. SubSplat achieves state-of-the-art performance across all metrics on RealEstate10K~\cite{zhou2018stereo} and ACID~\cite{liu2021infinite}, maintaining high fidelity with minimal degradation as the target resolution scales from $512 \times 512$ to $1024 \times 1024$ using the same $256 \times 256$ inputs. These results confirm that our approach remains robust against variations in rendering scale, consistently delivering high-quality reconstructions regardless of the target resolution.

\begin{table*}[t]
\setlength{\abovecaptionskip}{2pt}
\setlength{\belowcaptionskip}{1pt}
\renewcommand{\arraystretch}{1.1}
\setlength{\tabcolsep}{2pt} 

\caption{
Cost–performance comparison on RealEstate10K~\cite{zhou2018stereo} at a fixed target resolution. 
SubSplat achieves the best balance of quality and efficiency, delivering competitive reconstruction fidelity with significantly lower latency and peak memory by utilizing lower-resolution inputs.
}

\label{tab:cost_perf}
\centering
\resizebox{0.99\textwidth}{!}{
\begin{tabular}{l c c c c c c c}
\toprule
Method & Input Res. & Output Res. & Latency (ms) $\downarrow$  & Peak Memory (GB) $\downarrow$  & PSNR $\uparrow$ & SSIM $\uparrow$ & LPIPS $\downarrow$ \\
\midrule
PixelSplat~\cite{charatan2024pixelsplat} & $512 \times 512$ & $512 \times 512$ & 561 & 9.56 & 24.80 & 0.839 & 0.190 \\
MVSplat~\cite{chen2024mvsplat}           & $512 \times 512$ & $512 \times 512$ & 131 & 4.27 & 24.98 & 0.842 & 0.172 \\
TranSplat~\cite{zhang2025transplat}      & $512 \times 512$ & $512 \times 512$ & 192 & 5.14 & 23.50 & 0.804 & 0.197 \\
DepthSplat~\cite{xu2025depthsplat}       & $512 \times 512$ & $512 \times 512$ & \underline{123} & \underline{4.58} & 24.70 & \textbf{0.851} & 0.170 \\
HiSplat~\cite{tang2024hisplat}               & $512 \times 512$ & $512 \times 512$ & 1960 & 4.87 & \underline{25.34} & 0.850 & \underline{0.168} \\

\rowcolor{gray!15}
\textbf{Ours (SubSplat)}                 & $256 \times 256$ & $512 \times 512$ & \textbf{42} & \textbf{1.57} & \textbf{25.52} & \underline{0.850} & \textbf{0.167} \\
\bottomrule
\end{tabular}}
\end{table*}

\noindent\textbf{Efficiency and Cost-performance.}
We assess efficiency at a fixed $512 \times 512$ target by comparing pixel-aligned baselines against our configuration. While PixelSplat~\cite{charatan2024pixelsplat}, MVSplat~\cite{chen2024mvsplat}, and DepthSplat~\cite{xu2025depthsplat} deliver strong image quality, they incur significant latency and memory overhead due to their full-resolution input requirements. In particular, HiSplat~\cite{tang2024hisplat} achieves competitive reconstruction fidelity by extracting Gaussians across three distinct feature stages. However, this multi-stage feed-forward process leads to substantial latency degradation (1960ms), as shown in Table~\ref{tab:cost_perf}. In contrast, SubSplat achieves 25.52 PSNR and 0.850 SSIM at only 42ms per frame (24 FPS) by effectively restoring density from lower-resolution features. This makes it the only method among the evaluated baselines to achieve real-time performance. Note that baselines were not evaluated with $1024 \times 1024$ inputs due to Out-of-Memory (OOM) issues. These results demonstrate a superior cost-performance trade-off for our SPGR design, sustaining high-fidelity rendering with significantly lower computational and memory budgets.

\begin{figure}[t]
    \centering
    \includegraphics[width=1.0\linewidth]{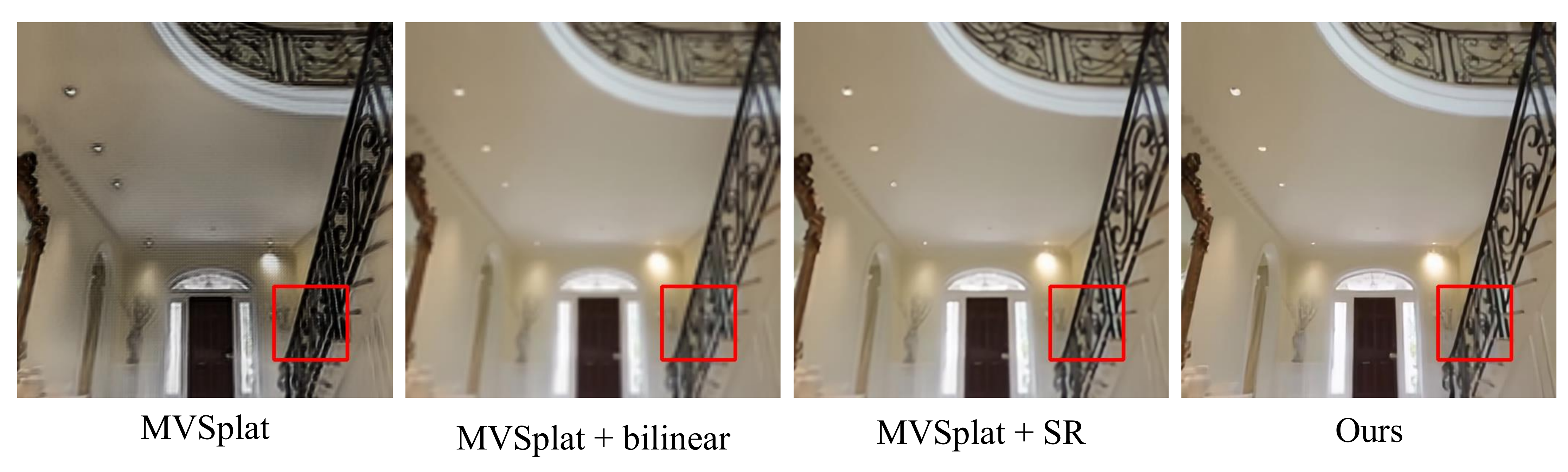}
    \vspace{-20pt}
    \caption{\textbf{Qualitative Comparison with Image-Space Upsamplers.} All methods render at $512 \times 512$ resolution starting from the same $256 \times 256$ inputs. While image-space upsamplers fail to recover structural details, SubSplat restores sharp textures and geometric consistency by generating high-density primitives directly in 3D space.}
    \label{fig:sr_comp}
    \end{figure}
\begin{table}[t]
\centering
\setlength{\tabcolsep}{13pt}
\renewcommand{\arraystretch}{1.1}

\caption{
\textbf{Comparison with image-space upsamplers on RealEstate10K~\cite{zhou2018stereo}.} 
All methods reconstruct from $256 \times 256$ inputs and render at $512 \times 512$ resolution. 
SubSplat significantly outperforms 2D post-processing methods by directly densifying geometry in 3D space, maintaining high fidelity with negligible latency overhead.
}
\vspace{-0.1cm}
\label{tab:upsampler_comparison} 
\resizebox{\columnwidth}{!}{
\begin{tabular}{l c c c c}
\toprule
Method & PSNR $\uparrow$ & SSIM $\uparrow$ & LPIPS $\downarrow$ & Latency $\downarrow$ (s) \\
\midrule
MVSplat + Bilinear (post)                & 24.01 & 0.801 & 0.237 & \textbf{0.042} \\
MVSplat + HiT-SR~\cite{zhang2024hit} (post) & 23.88 & 0.807 & 0.203 & 0.215 \\
\rowcolor{gray!15}
\textbf{Ours (SubSplat)}                 & \textbf{25.52} & \textbf{0.850} & \textbf{0.167} & \textbf{0.042} \\
\bottomrule
\end{tabular}}
\end{table}
    
\noindent\textbf{Comparison with Image-space Upsamplers.} We compare our approach against image-space upsamplers, including Bilinear interpolation and a $\times 2$ Super Resolution (SR) head~\cite{zhang2024hit}. While Bilinear interpolation remains blurry and the SR head introduces ringing and cross-view inconsistencies, SubSplat maintains structural clarity at $512 \times 512$ resolution. As 2D upsamplers operate purely in image space, they cannot resolve the underlying grid-anchored primitive limitations. In contrast, SubSplat optimizes Gaussian distributions via sub-pixel reparameterization, capturing geometry-aware details that image-space methods fail to recover. Quantitatively  (Table~\ref{tab:upsampler_comparison}), post-processing upsamplers are inferior to our configuration. SubSplat yields higher PSNR/SSIM and lower LPIPS with latency on par with Bilinear interpolation, while being significantly faster than the SR head. This demonstrates that image-space upsampling cannot compensate for under-resolved primitives, whereas our method delivers cleaner and more temporally stable results without additional post-processing.

\begin{table}[t]
\centering
\caption{
\textbf{Ablation Study on RealEstate10K.} We evaluate the effectiveness of each module at $\times 2$ ($K=4$) and $\times 4$ ($K=16$) scales. SPGR provides the primary performance gain, while Feature Aggregation ensures structural consistency as the scale increases.
}
\label{tab:integrated_ablation}
\setlength{\tabcolsep}{5pt}
\renewcommand{\arraystretch}{1.1}
\resizebox{\linewidth}{!}{
\begin{tabular}{l ccc ccc}
\toprule
\multirow{2}{*}{Modules} & \multicolumn{3}{c}{\textbf{$K=4$ ($512 \times 512$)}} & \multicolumn{3}{c}{\textbf{$K=16$ ($1024 \times 1024$)}} \\
\cmidrule(lr){2-4} \cmidrule(lr){5-7}
& PSNR $\uparrow$ & SSIM $\uparrow$ & LPIPS $\downarrow$ & PSNR $\uparrow$ & SSIM $\uparrow$ & LPIPS $\downarrow$ \\
\midrule
Baseline (MVSplat~\cite{chen2024mvsplat}) & 19.46 & 0.751 & 0.269 & 16.98 & 0.671 & 0.424 \\
+ Sub-pixel Reparam. & 25.15 & 0.843 & 0.173 & 19.49 & 0.736 & 0.415 \\
\rowcolor{gray!15} 
+ Feature Aggregation (Full) & \textbf{25.52} & \textbf{0.850} & \textbf{0.167} & \textbf{22.65} & \textbf{0.781} & \textbf{0.268} \\
\bottomrule
\end{tabular}}
\end{table}

\begin{table}[h]
\centering
\setlength{\tabcolsep}{15pt}
\renewcommand{\arraystretch}{1.1}

\caption{
\textbf{Ablation of Deformable Attention (DA) stages on RealEstate10K~\cite{zhou2018stereo}.} 
We evaluate the impact of incremental DA stages on scene reconstruction quality under the $256 \times 256$ to $512 \times 512$ configuration. 
Increasing the attention stages consistently improves structural fidelity and perceptual quality.
}
\vspace{-0.2cm}
\label{tab:da_stage_ablation}

\resizebox{0.85\columnwidth}{!}{
\begin{tabular}{l ccc}
\toprule
Configuration & PSNR $\uparrow$ & SSIM $\uparrow$ & LPIPS $\downarrow$ \\
\midrule
One-stage DA    & 23.94 & 0.833 & 0.194 \\
Two-stage DA    & 24.52 & 0.839 & 0.183 \\
\rowcolor{gray!15}
\textbf{Three-stage DA (Full)} & \textbf{25.52} & \textbf{0.850} & \textbf{0.167} \\
\bottomrule
\end{tabular}}
\vspace{-2mm}
\end{table}

\begin{figure}[t!]
\centering
\includegraphics[width=\linewidth]{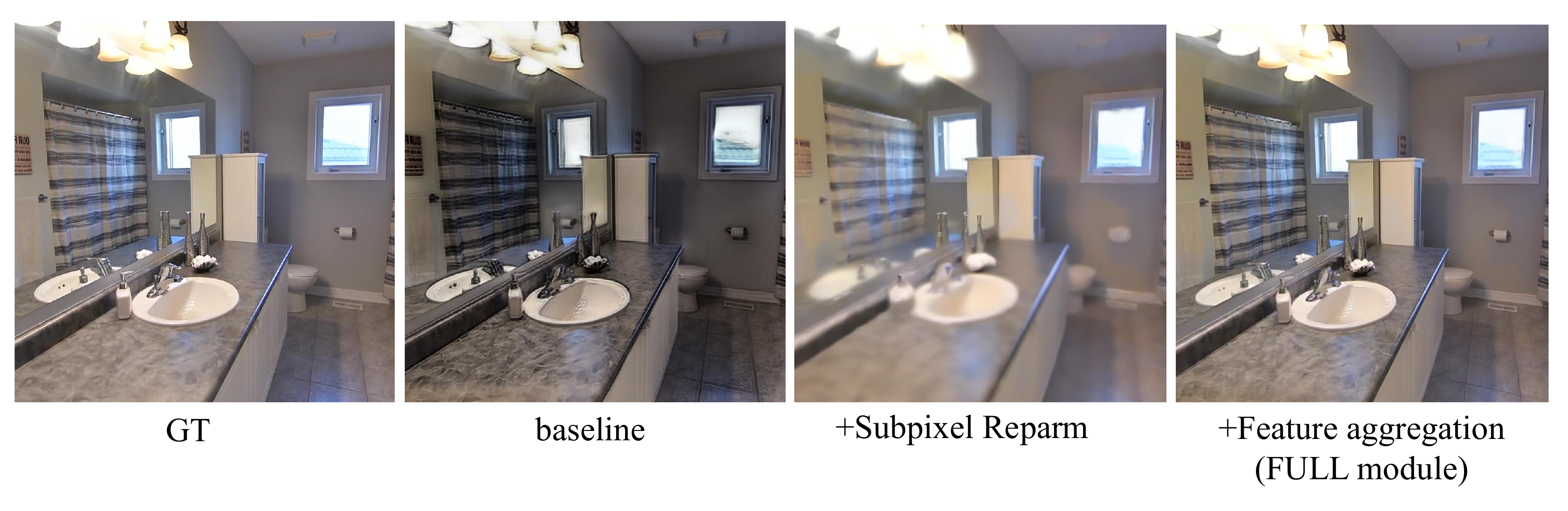}
\vspace{-0.8cm}
\caption{
\textbf{Ablation study of SubSplat modules across subdivision scales.} We evaluate the effectiveness of each module. SPGR provides the primary performance gain, while Feature Aggregation ensures structural consistency as the scale increases.
}
\vspace{-0.4cm}
\label{fig:ablation_study}
\end{figure}

\subsection{Ablation Study}
We evaluate the contribution of SubSplat modules across subdivision scales $\times 2$ ($K=4$) and $\times 4$ ($K=16$). While SPGR acts as the primary recovery mechanism by providing a +5.69 dB PSNR boost at $K=4$, the significance of Feature Aggregation grows with task difficulty, increasing its contribution to +3.16 dB at the $K=16$. As illustrated in Fig.~\ref{fig:ablation_study}, this synergy is effective for restoring high-frequency textures and preserving structural clarity at larger scales. Additionally, Table~\ref{tab:da_stage_ablation} shows the impact of Deformable Attention (DA) stages. Performance improves with more stages as iterative feature sampling better captures spatial details, where a three-stage configuration yields a 1.58 dB PSNR boost over the one-stage baseline.
\section{Conclusion}
We have proposed SubSplat, a novel framework that achieves high-fidelity rendering by generating fine-grained sub-pixel primitives directly from low-resolution features. This mechanism mitigates the quadratic cost of backbones with high-resolution inputs while maintaining stable network latency across extended output scales. Although rendering overhead increases with primitive count, SubSplat delivers a superior quality-efficiency trade-off and real-time performance. Future work will explore content-aware primitive density control to further optimize the balance between rendering throughput and structural fidelity.

\section*{Acknowledgements}
This research was supported by the Basic Science Research Program of the National Research Foundation (NRF) funded by the Korean government (MSIT) (No. IITP-2026-RS-2024-00346737), by the NRF grant funded by the Korea government(MSIT) (No. RS-2025-00563942), by the IITP grant funded by the Korea government(MSIT)(No. IITP-2026-RS-2020-II201819, 10\%), and by the AI Computing Infrastructure Enhancement (GPU Rental Support) User Support Program funded by the Ministry of Science and ICT (MSIT), Republic of Korea.

%
%
\bibliographystyle{splncs04}
\bibliography{main}

\clearpage
\appendix
\authorrunning{Jiun Lee et al.}
\titlerunning{SubSplat}
\title{SubSplat: High-Resolution Pixel-aligned 3DGS via Sub-pixel Gaussian Reparameterization \\[0.5em] \normalfont{Supplementary Material}} 
\author{}
\institute{}
\maketitle

\setcounter{figure}{0}
\renewcommand{\thefigure}{\Alph{figure}}

\setcounter{table}{0}
\renewcommand{\thetable}{\Alph{table}}
We provide additional technical details and extended results that complement the main paper. Appendix~\hyperref[sec:supp_A]{A} describes the network architecture, covering the pixel-aligned backbone, query embeddings, Deformable Attention in Feature aggregation, and the Sub-pixel Gaussian Reparameterizer (SPGR) formulation, supplemented by ablation studies on key hyperparameters. Appendix~\hyperref[sec:supp_B]{B} presents supplementary quantitative and qualitative results at $512\times512$ and $1024\times1024$ resolutions from a fixed $256\times256$ input, alongside comparisons with uniform density scaling and post-processing upsamplers. Additionally, a video demonstration is provided.  Appendix~\hyperref[sec:supp_C]{C} discusses the limitations of our approach and suggests directions for future work.

\section*{A. Network Architecture Details}
\label{sec:supp_A}
\noindent\textbf{A.1\quad Backbone and Query Embeddings} 
\label{sec:supp_A1}
We employ the pixel-aligned backbone of MVSplat~\cite{chen2024mvsplat} to extract multi-view feature maps. To facilitate the subsequent sub-pixel reparameterization, each primary Gaussian state $a_p=[\mu_p, s_p, q_p, \alpha_p]$ is encoded into a high-dimensional feature space using four dedicated MLPs: $\phi_x$ for position, $\phi_s$ for scale, $\phi_q$ for quaternion, and $\phi_\alpha$ for opacity. Each MLP produces a 128-dimensional embedding, forming the geometric basis for sub-pixel Gaussian reparameterization. The corresponding primary SH color (75-dimensional when flattened) is processed by another MLP $\phi_c$, which also yields a 128-dimensional embedding to guide appearance redistribution. Consequently, all geometry and appearance queries are aligned to a shared embedding dimension $E=128$, providing a consistent contextual input for the SPGR module.

\subsection*{A.2\quad Deformable Attention}
\label{sec:supp_A2}
Our deformable-attention operator follows the formulation used in 
GaussianFormer~\cite{huang2024gaussianformer} and 
PixelGaussian~\cite{fei2024pixelgaussian}, and we adopt the PixelGaussian
sampling setup for aggregating multi-view evidence.
The backbone provides multi-view feature maps 
$\mathcal{F}=\{\Phi_{v,\ell}\}$ with 
$\Phi_{v,\ell}\in\mathbb{R}^{H_\ell\times W_\ell\times D_\ell}$.
Each primary Gaussian consists of geometry attributes 
$a_p=[\mu_p,s_p,q_p,\alpha_p]$ and a degree-4 SH color vector 
$h_p\in\mathbb{R}^{75}$, which are encoded into 
$z_p^{g}, z_p^{a}\in\mathbb{R}^{E}$ ($E{=}128$).

\begin{equation}
    R_v = P(\mu_p).
\end{equation}

Each anchor location $\mu_p$ is projected into view $v$ using dataset intrinsics
and extrinsics to obtain the reference point $R_v$.

\begin{equation}
\phi(\Phi_{v,\ell}, R_v)
=
\sum_{(i,j)}
g(R_{v,x}, i)\, g(R_{v,y}, j)\, \Phi_{v,\ell}(i,j),
\end{equation}

$\phi(\cdot)$ extracts a localized descriptor at $R_v$ via bilinear weight $g(a,b)=\max(0,1-|a-b|)$.

\begin{equation}
    W_v(f,z)
    =
    \mathrm{softmax}\!\left(
        (f + z)\cdot \phi(\Phi_{v,\ell},R_v)^{\!\top}
    \right),
\end{equation}

Here, $f$ denotes either a geometry or appearance feature, depending on the DA stage,
and $z$ is the corresponding geometry or appearance embedding.

\begin{equation}
\mathrm{DA}(\mathcal{F}; z,f,P(\mu_p),a_p)
=\sum_{v=1}^{V}\alpha_v\, W_v(f,z)\,\phi(\Phi_{v,\ell},R_v).
\end{equation}

The deformable-attention operator aggregates view-specific evidence through the view gates $\alpha_v$, returning an updated query descriptor. Consistent with the setup in PixelGaussian~\cite{fei2024pixelgaussian}, these view gates allow the model to weight the importance of each viewpoint during feature aggregation. In practice, geometry applies deformable attention three times, while appearance applies it once.

\subsection*{A.3\quad Sub-pixel Gaussian Reparameterizer}
\label{sec:supp_A3}
The Sub-pixel Gaussian Reparameterizer converts a primary Gaussian $a_p=[\mu_p,s_p,q_p,\alpha_p]$
into $K$ sub-pixel Gaussians by refining its position, depth, scale,
rotation, and opacity. Our formulation follows the update rules in Eq.~(9)--(13)
of the main paper and matches our implementation.

\noindent\textbf{Primary-view selection.}
For each primary Gaussian anchor $\mu_p$, we project it into all cameras and choose the
primary view $v^\ast$ as the one with the smallest positive depth:
\begin{equation}
R_v = P(\mu_p), 
\qquad 
v^\ast = \arg\min_{v:\,z_v>0} z_v .
\end{equation}
This yields stable back-projection and consistent reparameterization.

\noindent\textbf{Primary-view projection Jacobian.}
We estimate each sub-pixel Gaussian’s screen-space footprint using the projection Jacobian
of the primary view $v^\ast$, evaluated at $\mu_{p,k}$, following 3DGS~\cite{kerbl20233d}. Let $(f_x,f_y)$ be pixel-space focal lengths (scaled to the render
resolution), and let $(x_c,y_c,z_c)$ denote the camera-frame coordinates at
$\mu_{p,k}$ in the primary view. The pinhole Jacobian w.r.t.\ camera coordinates is
\begin{equation}
B =
\begin{bmatrix}
f_x/z_c & 0         & -f_x\,x_c/z_c^2\\
0       & f_y/z_c   & -f_y\,y_c/z_c^2
\end{bmatrix},
\end{equation}
and the world-space projection Jacobian (chain rule) is
\begin{equation}
J_{v^\ast}(\mu_{p,k}) \;=\; B\,R^{v^\ast}_{\mathrm{w2c}}.
\end{equation}
The footprint area used in opacity allocation is approximated by
\begin{equation}
A_k \;\approx\; \sqrt{\det\!\big(J_{v^\ast}(\mu_{p,k})\,\Sigma_{p,k}\,J_{v^\ast}(\mu_{p,k})^\top\big)}.
\end{equation}

\noindent\textbf{Bounded depth residual.}
Let $R_{\mathrm{c2w}}$ be the camera-to-world rotation of the primary view, and
let $\mathbf{b}_u,\mathbf{b}_v$ denote the world-space lateral ray bases
corresponding to one-pixel shifts at the primary Gaussian projection (computed from the
primary intrinsics/extrinsics). Let $\mathbf{b}_z = R_{\mathrm{c2w}}\,[0,0,1]^\top$
be the forward ray direction. We update along-ray motion with a bounded residual:
\begin{equation}
\Delta\hat z_k \;=\; \beta_z\,\tanh(\Delta z_k).
\end{equation}
The child center is then
\begin{equation}
\mu_{p,k}
\;=\;
\mu_p
\;+\;
\Delta u_k\,\mathbf{b}_u
\;+\;
\Delta v_k\,\mathbf{b}_v
\;+\;
\Delta\hat z_k\,\mathbf{b}_z.
\end{equation}
We found $\beta_z{=}0.25$ to offer the best PSNR/SSIM/LPIPS trade-off 
\begin{table}[t]
\centering
\caption{\textbf{Ablation on depth–residual scale $\beta_z$.}
Best values in each metric are highlighted.}
\vspace{3pt}
\begin{tabular}{cccc}
\toprule
$\beta_z$ & PSNR $\uparrow$ & SSIM $\uparrow$ & LPIPS $\downarrow$ \\
\midrule
0.50        & 24.92 & 0.846 & 0.169 \\
0.75        & 24.36 & 0.841 & 0.173 \\
1.00        & 24.14 & 0.835 & 0.176 \\
\rowcolor{gray!10} 
\textbf{0.25 (ours)} & \textbf{25.52} & \textbf{0.850} & \textbf{0.167} \\
\bottomrule
\end{tabular}
\label{tab:depth_beta}
\end{table}

\begin{table}[t]
\centering
\caption{\textbf{Ablation on scale factor $\lambda_s$.} Best values in each metric are highlighted.}
\vspace{3pt}
\begin{tabular}{c ccc}
\toprule
Scale ($\lambda_s$) & PSNR $\uparrow$ & SSIM $\uparrow$ & LPIPS $\downarrow$ \\
\midrule
0.50 & 23.22 & 0.831 & 0.208 \\
0.55 & 24.81 & 0.844 & 0.176 \\
0.65 & 25.12 & 0.848 & 0.174 \\
0.70 & 24.39 & 0.842 & 0.184 \\

\rowcolor{gray!10} 
\textbf{0.60 (Ours)} & \textbf{25.52} & \textbf{0.850} & \textbf{0.167} \\
\bottomrule
\end{tabular}
\label{tab:ablation_scale}
\end{table}

\noindent\textbf{Scale modulation}
Not included in the main paper for simplicity, the scale factor $\lambda_s$ is adjusted to support adequate spatial coverage during sub-pixel Gaussian division. This value is intended to maintain structural continuity by mitigating potential gaps between subdivided Gaussians. Specifically, the scale factor follows the heuristic $\lambda_s \approx 1/\sqrt{K}$, which corresponds to the side length of each sub-region when a unit pixel area is partitioned into $K$ sub-pixels. We found that $\lambda_s = 0.60$ provides a well-balanced distribution and the best performance (Table~\ref{tab:ablation_scale}), and this configuration was used for all results reported in the main text.

\noindent\textbf{Color modulation (bounded gain).}
To stabilize appearance updates, we apply a scalar gain factor $\gamma_k$ to the
primary Gaussian SH color $h_p\!\in\!\mathbb{R}^{75}$ when forming the sub-pixel Gaussian color:
\begin{equation}
h_{p,k} \;=\; \gamma_k \, h_p .
\end{equation}
The network predicts an unconstrained logit $\tilde{\gamma}_k\!\in\!\mathbb{R}$, which
we squash into a bounded interval around $1$:
\begin{equation}
\gamma_k \;=\; 1 \;+\; \delta\,\tanh(\tilde{\gamma}_k),
\qquad \delta \in [0,1].
\end{equation}
This symmetric mapping guarantees $\gamma_k \in [\,1-\delta,\;1+\delta\,]$, preventing  
over‑/under‑ amplification while keeping gradients well‑behaved. We sweep $\delta$ to realize the intervals $[0.5,1.5]$ ($\delta{=}0.5$),
$[0.75,1.25]$ ($\delta{=}0.25$), and the no‑gain case $[1.0,1.0]$ ($\delta{=}0$).
A moderate bound $[0.75,1.25]$ (i.e., $\delta{=}0.25$) offered the best
PSNR/SSIM /LPIPS trade‑off in our experiments (Table~\ref{tab:color_gain}), whereas a wider
range induced color drift and the no‑gain setting reduced local adaptability.
\begin{table}[t]
\centering
\caption{\textbf{Ablation on bounded color gain $\gamma_k$.}
Best values are highlighted.}
\vspace{3pt}
\begin{tabular}{cccc}
\toprule
Gain range & PSNR $\uparrow$ & SSIM $\uparrow$ & LPIPS $\downarrow$ \\
\midrule
$[0.5,\,1.5]$      & 24.71  & 0.843  & 0.171 \\
$[1.0,\,1.0]$ & 24.89 & 0.844 & 0.170 \\
\rowcolor{gray!10} 
$[0.75,\,1.25]$ (ours) & \textbf{25.52} & \textbf{0.850} & \textbf{0.167} \\
\bottomrule
\end{tabular}
\label{tab:color_gain}
\end{table}

\noindent\textbf{Anti-aliasing via minimum footprint.}
Following EWA Splatting~\cite{zwicker2002ewa} and Mip-Splatting~\cite{yu2024mip},
we keep each projected Gaussian at least at the pixel scale by clamping
its screen-space area:
\begin{equation}
A_k \leftarrow \max(A_k,\,A_{\min}).
\end{equation}
To prevent sub-pixel collapse during high-resolution reparameterization, we scale $A_{\min}$ according to the subdivision factor $K$. We set $A_{\min}=0.25$ ($1/K$) for $K=4$ subdivision and $A_{\min}=0.0625$ ($1/K$) for $K=16$ subdivision in units of the target pixel scale. This ensures that each generated primitive maintains sufficient coverage to act as a robust anti-aliasing safeguard while preserving structural continuity across different reparameterization scales.

\begin{table}[t]
\centering
\caption{\textbf{Quantitative comparison of uniform density scaling vs. SubSplat on RealEstate10K~\cite{zhou2018stereo}.} 
All methods use a $256\times256$ input resolution. The baseline MVSplat~\cite{chen2024mvsplat} is modified to generate $K$ Gaussians as SubSplat to match the target resolutions ($512\times512$ and $1024\times1024$). 
Metrics: PSNR$\uparrow$ / SSIM$\uparrow$ / LPIPS$\downarrow$.}
\vspace{3pt}
\begin{tabular}{l ccc ccc}
\toprule
\multirow{2}{*}{Method} & \multicolumn{3}{c}{$K=4$ ($512\times512$)} & \multicolumn{3}{c}{$K=16$ ($1024\times1024$)} \\
\cmidrule(lr){2-4} \cmidrule(lr){5-7}
& PSNR $\uparrow$ & SSIM $\uparrow$ & LPIPS $\downarrow$ & PSNR $\uparrow$ & SSIM $\uparrow$ & LPIPS $\downarrow$ \\
\midrule
MVSplat~\cite{chen2024mvsplat} (baseline) & 20.13 & 0.711 & 0.320 & 19.42 & 0.668 & 0.455 \\
\rowcolor{gray!10} 
\textbf{SubSplat (Ours)} & \textbf{25.52} & \textbf{0.850} & \textbf{0.167} & \textbf{22.65} & \textbf{0.781} & \textbf{0.268} \\
\bottomrule
\end{tabular}
\label{tab:scaling_comparison}
\end{table}

\section*{B. Additional Results}
\label{sec:supp_B}
\subsection*{B.1\quad Analysis on Uniform Density Scaling vs. Sub-pixel Gaussian Reparameterization.} 
\label{sec:supp_B1}
To verify that high-resolution rendering requires more than a simple increase in primitive density, we compare SubSplat against our baseline, MVSplat~\cite{chen2024mvsplat}, on RealEstate10K~\cite{zhou2018stereo}. From a $256 \times 256$ input, we evaluate performance at target resolutions of $512 \times 512$ ($K=4$) and $1024 \times 1024$ ($K=16$). For a comparison, the baseline is modified to generate the same $K$ Gaussians as our subdivision scale. As shown in Table~\ref{tab:scaling_comparison}, simply increasing the baseline's primitive count yields significantly lower performance, as Gaussians generated without sub-pixel guidance lead to redundant representations. In contrast, SubSplat’s SPGR breaks this resolution bottleneck by computing distinct sub-pixel offsets and attribute modulations. These results confirm that reparameterization, rather than a quantitative expansion of the baseline, is critical for overcoming the backbone’s resolution limit.

\begin{table}[t]
\centering
\caption{\textbf{Efficiency analysis on RealEstate10K~\cite{zhou2018stereo}.} 
We compare the cost of different input resolutions for $512 \times 512$ output. 
Latency includes both network processing and rendering times.}
\vspace{3pt}
\begin{tabular}{l cc cc}
\toprule
\multirow{2}{*}{Method} & \multicolumn{2}{c}{Input Res. $256 \times 256$} & \multicolumn{2}{c}{Input Res. $512 \times 512$} \\
\cmidrule(lr){2-3} \cmidrule(lr){4-5}
& Latency (s) & Mem (GB) & Latency (s) & Mem (GB) \\
\midrule
PixelSplat~\cite{charatan2024pixelsplat} & 0.137 & 2.84 & 0.561 & 9.56 \\
TranSplat~\cite{zhang2025transplat} & 0.063 & 2.78 & 0.192 & 5.14 \\
MVSplat~\cite{chen2024mvsplat} & 0.041 & 1.20 & 0.131 & 4.27 \\
HiSplat~\cite{tang2024hisplat} & 0.490 & 2.33 & 1.960 & 4.87 \\
DepthSplat~\cite{xu2025depthsplat} & 0.038 & 2.62 & 0.123 & 4.58 \\
\rowcolor{gray!10} 
\textbf{SubSplat (Ours)} & \textbf{0.042} & \textbf{1.57} & - & - \\
\bottomrule
\end{tabular}
\label{tab:latency}
\end{table}

\subsection*{B.2\quad Computational Efficiency Analysis}
\label{sec:supp_B2}
Increasing the backbone resolution in MVSplat~\cite{chen2024mvsplat} significantly raises latency and memory consumption, making high-resolution inference impractical. As shown in Table~\ref{tab:latency}, other baseline architectures also suffer from a severe escalation in latency when the input resolution is increased, as the computational complexity of their feature extractors and attention mechanisms scales quadratically. SubSplat addresses this by achieving high-resolution output with a $256 \times 256$ input-resolution backbone. By utilizing the Sub-pixel Gaussian Reparameterizer (SPGR), SubSplat decouples output fidelity from the backbone's load, resulting in lower latency and memory usage than the baseline. This confirms that reparameterization is a more reasonable solution than simple input resolution expansion for restoring structural details without prohibitive computational overhead.

\begin{table}[t]
\centering
\caption{
\textbf{Quantitative comparison of post-processing upsamplers on RealEstate10K~\cite{zhou2018stereo}.} 
All methods generate a $1024 \times 1024$ output from a $256 \times 256$ input. 
SubSplat outperforms MVSplat~\cite{chen2024mvsplat} combined with bilinear or HiT-SR~\cite{zhang2024hit} upsampling in both quality and efficiency. 
Metrics: PSNR$\uparrow$, SSIM$\uparrow$, LPIPS$\downarrow$, and Latency (s)$\downarrow$.
}
\vspace{-0.3cm}
\setlength{\tabcolsep}{4pt} 
\begin{tabular}{p{3.0cm}cccc}
\toprule
Method & PSNR$\uparrow$ & SSIM$\uparrow$ & LPIPS$\downarrow$ & Latency \\
\midrule
MVSplat + bilinear      & 21.46   & 0.741   & 0.316  & 0.070  \\
MVSplat + HiT-SR         & 22.37   & 0.774   & 0.293   & 0.132  \\
\rowcolor{gray!10} 
\textbf{Ours}           & \textbf{22.65} & \textbf{0.781} & \textbf{0.268} & \textbf{0.045} \\
\bottomrule
\end{tabular}
\label{tab:upsampling_256_1024}

\end{table}

\begin{figure*}[b]
\vspace{-1cm}
\centering
\includegraphics[width=\textwidth]{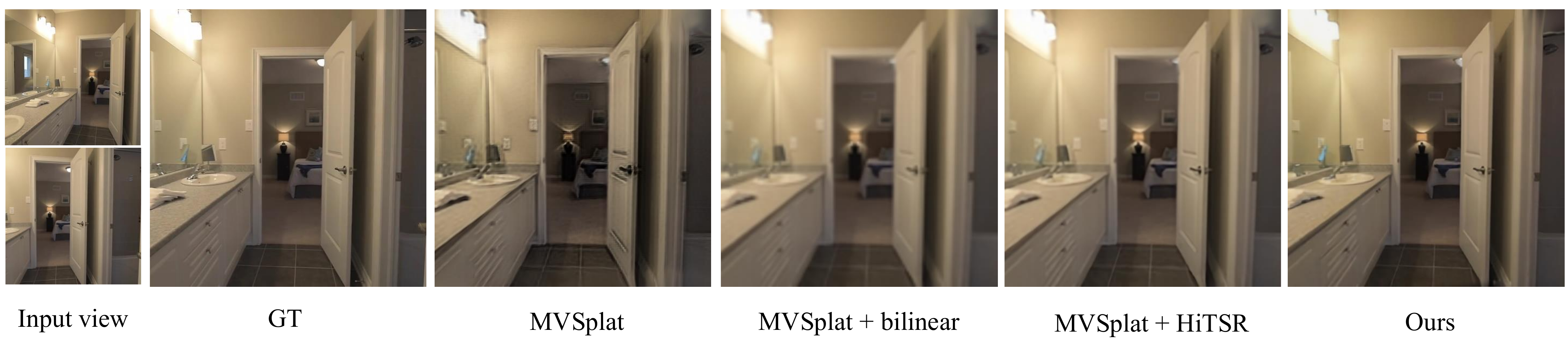}
\caption{
\textbf{Additional qualitative results on RealEstate10K~\cite{zhou2018stereo} for upsampling.} 
All results compare $1024 \times 1024$ output resolutions generated from a $256 \times 256$ input.
}
\label{fig:appendix_upsampling_1024}

\end{figure*}
\begin{figure*}[b!]
  \centering
 
  \includegraphics[width=\textwidth]{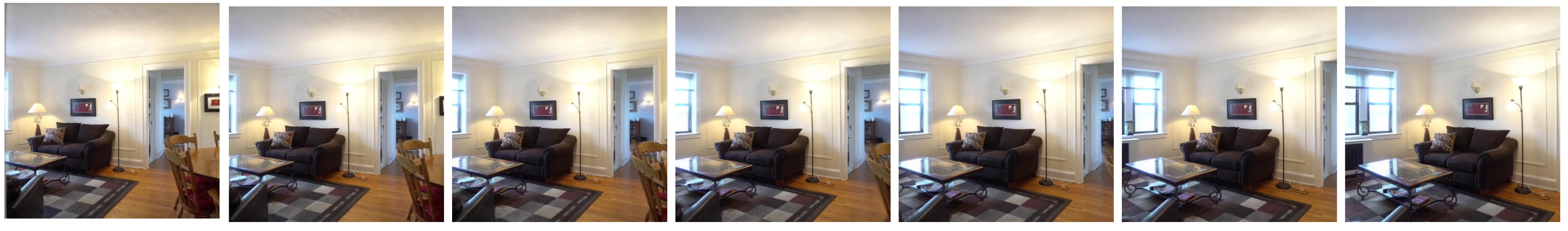}
  \vspace{-7mm}
  \caption{\textbf{Video demonstration} Upscaled 120-frame sequence (256$\times$256 to 512$\times$512) demonstrating temporal consistency.}
  \label{fig:video_demo}
\end{figure*}

\subsection*{B.3 Additional Comparison}
\label{sec:supp_B3}
\begin{figure*}[t!]
\centering

\includegraphics[width=\textwidth]{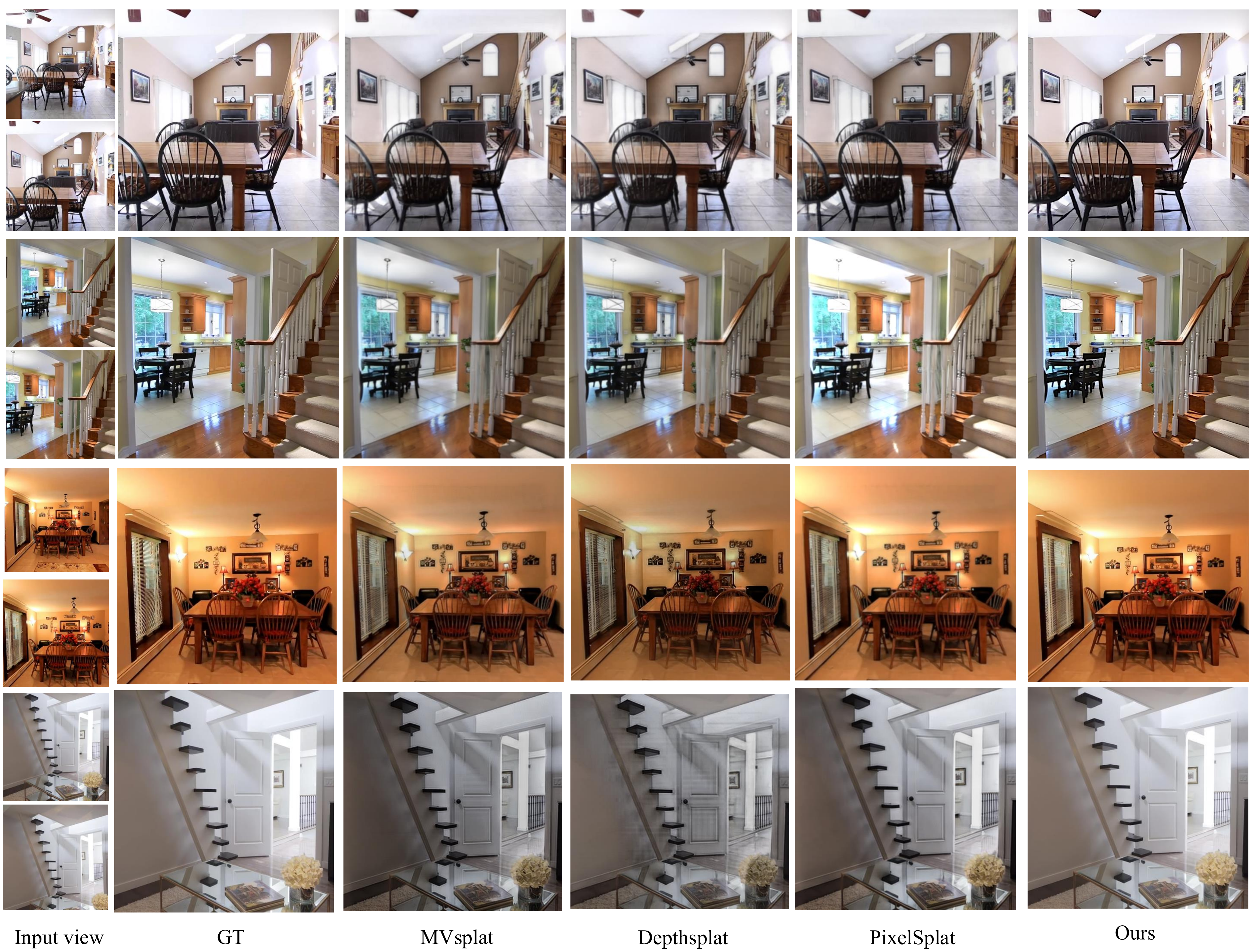}

\includegraphics[width=\textwidth]{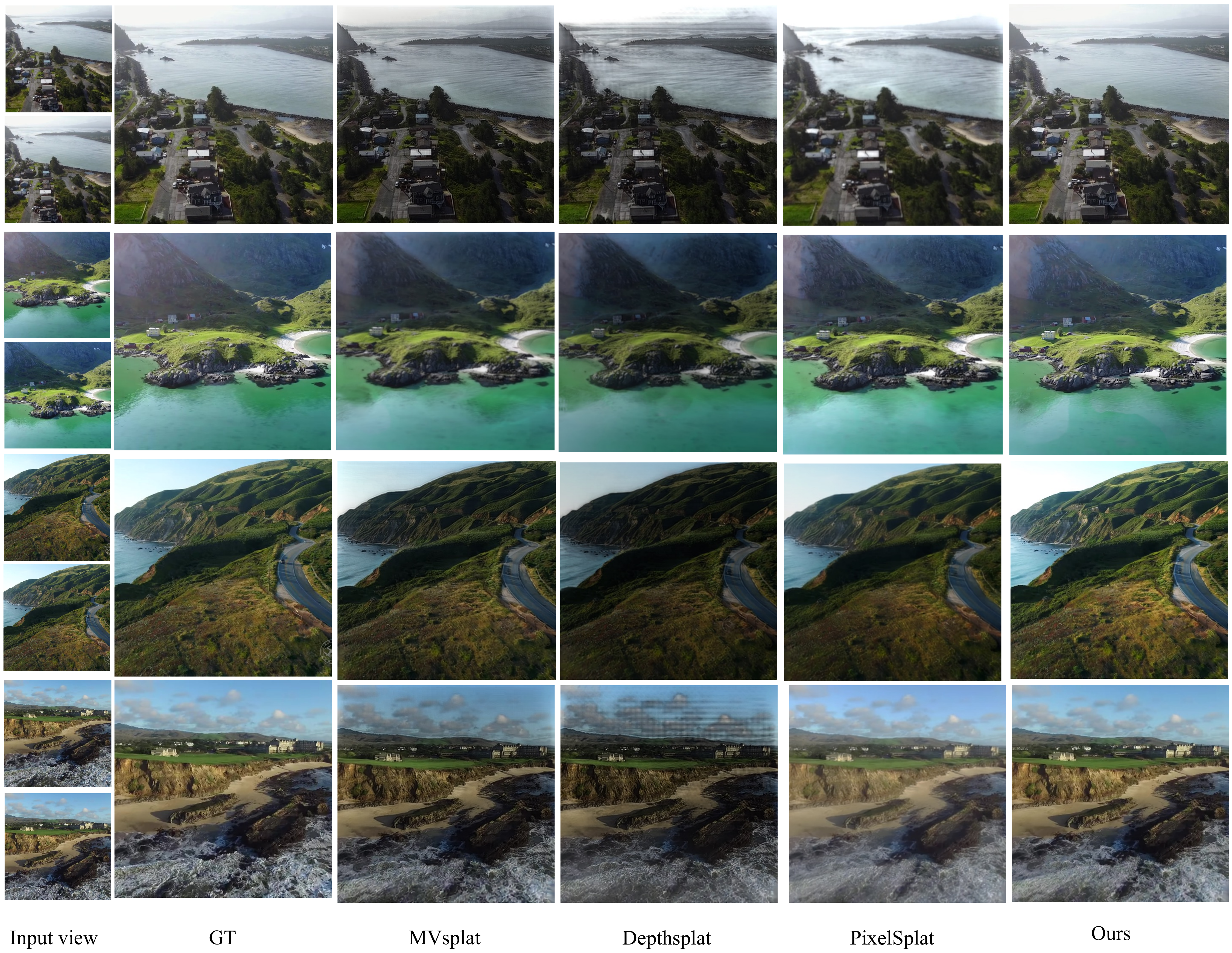}
\vspace{-10pt}
\caption{
\textbf{Additional qualitative results on RealEstate10K~\cite{zhou2018stereo} and ACID~\cite{liu2021infinite}.} 
All results are generated at a $512 \times 512$ output resolution from a $256 \times 256$ input.
}
\label{fig:appendix_256_512_all}
\end{figure*}

\begin{figure*}[t]
\centering
\includegraphics[width=\textwidth]{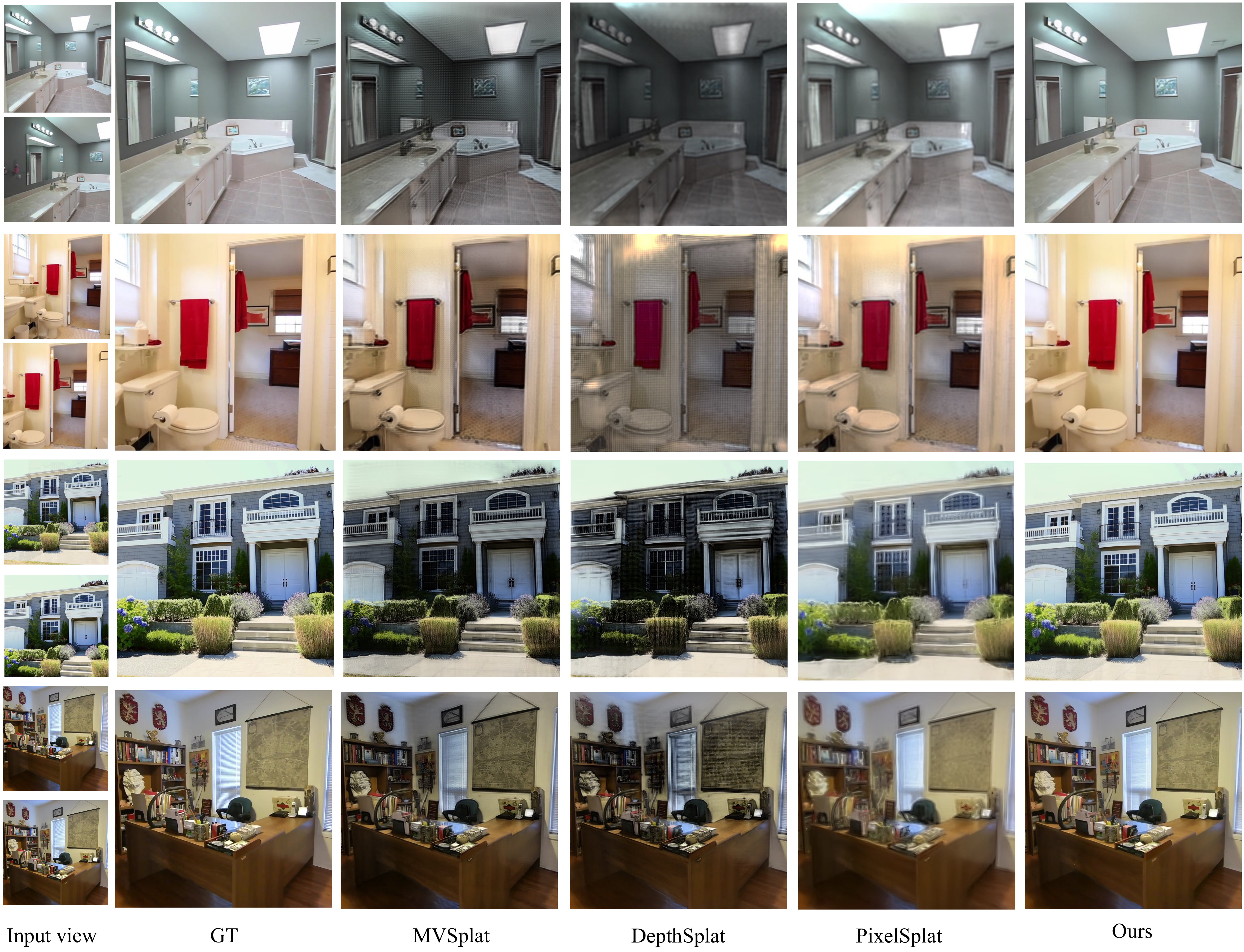}

\includegraphics[width=\textwidth]{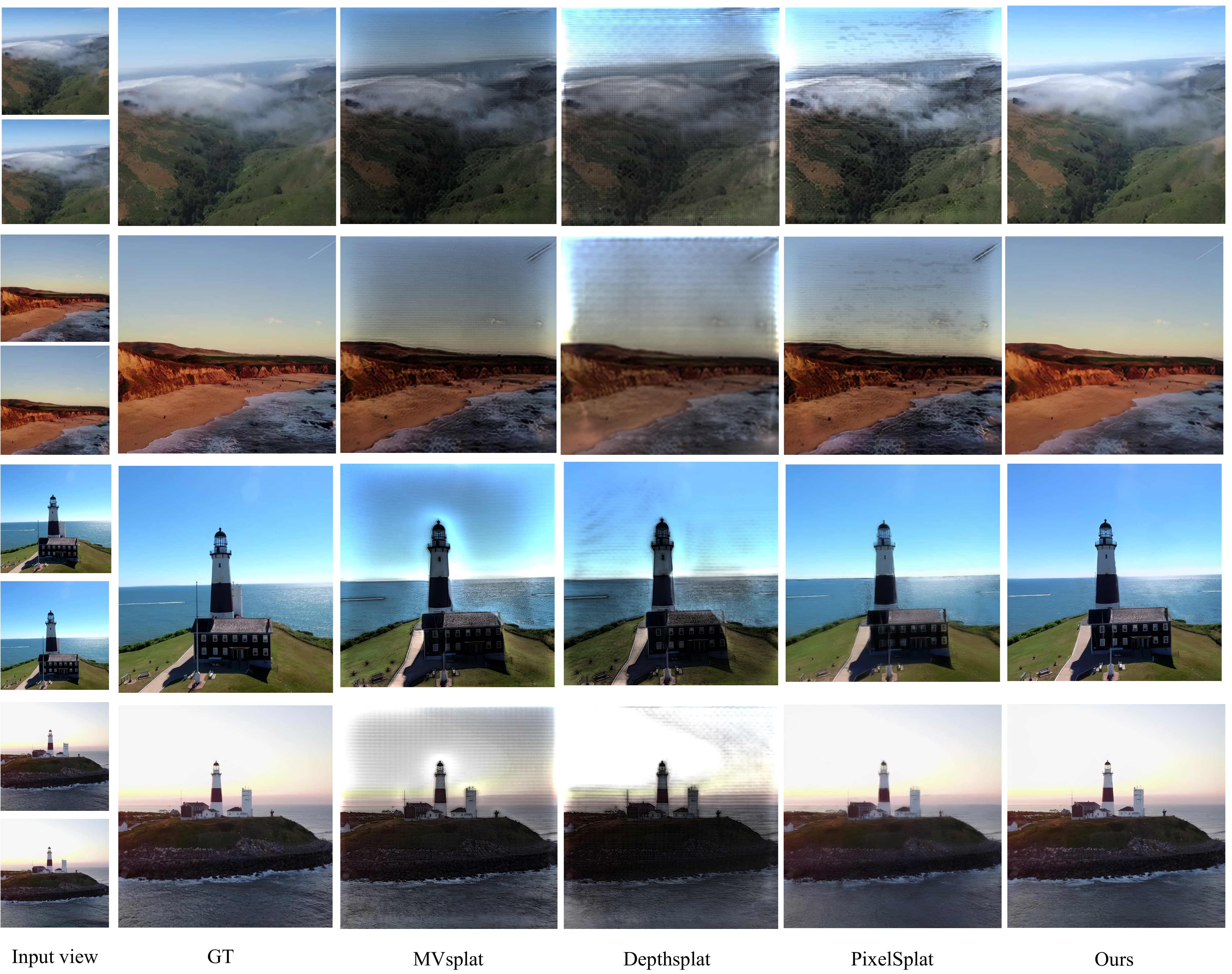}
\vspace{-10pt}
\caption{
\textbf{Additional qualitative results on RealEstate10K~\cite{zhou2018stereo} and ACID~\cite{liu2021infinite}.} 
All results are generated at a $1024 \times 1024$ output resolution from a $256 \times 256$ input.
}
\label{fig:appendix_256_1024_all}
\end{figure*}

\vspace{-0.3cm}
We provide additional quantitative results on the RealEstate10K~\cite{zhou2018stereo} and ACID~\cite{liu2021infinite} datasets, evaluating the performance of SubSplat at $512 \times 512$ and $1024 \times 1024$ output resolutions from a $256 \times 256$ input. Furthermore, we include a comparative analysis between our approach and baseline models paired with post-processing with 2D upsampling for the $1024 \times 1024$ output resolution, as reported in Table~\ref{tab:upsampling_256_1024}. Corresponding qualitative results are provided in Figure~\ref{fig:appendix_256_512_all}, Figure~\ref{fig:appendix_256_1024_all}, and Figure~\ref{fig:appendix_upsampling_1024}. Finally, Figure~\ref{fig:video_demo} presents a 120-frame sequence upscaled from $256 \times 256$ to $512 \times 512$ at 20-frame intervals, demonstrating fine temporal consistency.
\vspace{-14pt}

\section*{C.Limitations and Future Work}
\label{sec:supp_C}
\vspace{-0.3cm}

The current implementation of our method relies on a fixed subdivision factor $K$, which restricts its ability to adaptively handle arbitrary scaling factors or various target resolutions. Additionally, as our approach is built upon MVSplat~\cite{chen2024mvsplat}, it can be affected by extreme occlusion and large viewpoint variations, where decreased depth estimation accuracy may lead to geometric inconsistencies. Future research could explore adaptive subdivision mechanisms and more robust, occlusion-aware cost volumes to improve structural stability across diverse and complex scenes.

\begingroup  

\setlength{\textfloatsep}{4pt}
\setlength{\floatsep}{4pt}
\setlength{\dbltextfloatsep}{4pt}
\setlength{\dblfloatsep}{4pt}
\makeatletter
\setlength{\@fptop}{0pt}
\setlength{\@fpsep}{4pt}
\makeatother

\end{document}